\documentclass[10pt,journal,compsoc]{IEEEtran}
\ifCLASSOPTIONcompsoc
  \usepackage[nocompress]{cite}
\else
  \usepackage{cite}
\fi
\ifCLASSINFOpdf
\else
\fi

\usepackage[utf8]{inputenc} 
\usepackage[T1]{fontenc}    
\usepackage{hyperref}       
\usepackage{url}            
\usepackage{booktabs}       
\usepackage{amsfonts}       
\usepackage{nicefrac}       
\usepackage{microtype}      
\usepackage{xcolor}         
\usepackage{multirow}
\usepackage{wrapfig}

\usepackage{enumitem}
\setlist{wide,labelindent=0.5em,leftmargin=0.5em}

\usepackage{graphics} 
\usepackage{epsfig} 
\usepackage{mathptmx} 
\usepackage{times} 
\usepackage{amsmath} 
\usepackage{amssymb}  
\usepackage{xcolor}
\usepackage{bbding}
\usepackage{bbold}
\usepackage{hyperref}
\usepackage{booktabs}
\usepackage{graphicx}
\usepackage{subfigure}
\usepackage[noend]{algorithm2e}
\usepackage{algcompatible}%
\usepackage{amssymb}
\usepackage{pifont}
\newcommand{\cmark}{\ding{51}}%
\let\oldnl\nl
\newcommand{\nonl}{\renewcommand{\nl}{\let\nl\oldnl}}

\definecolor{LQY_color}{RGB}{50, 205, 50}

\newcommand{\codefragment}[1]{\textcolor{purple}{\tt #1}}

\definecolor{revision_color}{RGB}{255, 0, 0}

\newcommand{\revision}[1]{#1}  

\usepackage{float}

\usepackage{lipsum}

\ifCLASSOPTIONcompsoc
  \usepackage[nocompress]{cite}
\else
  \usepackage{cite}
\fi

\begin{document}

\title{
MetaDrive: Composing Diverse Driving Scenarios for Generalizable Reinforcement Learning
}

%

\author{%
Quanyi Li\textsuperscript{$\mathsection$*},
Zhenghao Peng\textsuperscript{$\dagger$*},
Lan Feng\textsuperscript{$\ddagger$},
Qihang Zhang\textsuperscript{$\dagger$}, 
Zhenghai Xue\textsuperscript{$\dagger$},
Bolei Zhou\textsuperscript{$\spadesuit$}\\ %
\textsuperscript{$\mathsection$}Centre for Perceptual and Interactive Intelligence, Hong Kong SAR, China,\\
\textsuperscript{$\dagger$}The Chinese University of Hong Kong, Hong Kong SAR, China,  
\textsuperscript{$\ddagger$}ETH Zurich, Switzerland, \\
\textsuperscript{$\spadesuit$}University of California, Los Angeles, USA.
\thanks{\textsuperscript{*}Quanyi Li and Zhenghao Peng contribute equally to this work.}
}


\IEEEtitleabstractindextext{%
\begin{abstract}
Driving safely requires multiple capabilities from human and intelligent agents, such as the generalizability to unseen environments, the safety awareness of the surrounding traffic, and the decision-making in complex multi-agent settings. Despite the great success of Reinforcement Learning (RL), most of the RL research works investigate each capability separately due to the lack of integrated environments. In this work, we develop a new driving simulation platform called MetaDrive to support the research of generalizable reinforcement learning algorithms for machine autonomy. MetaDrive is highly compositional, which can generate an infinite number of diverse driving scenarios from both the procedural generation and the real data importing. Based on MetaDrive, we construct a variety of RL tasks and baselines in both single-agent and multi-agent settings, including benchmarking generalizability across unseen scenes, safe exploration, and learning multi-agent traffic. The generalization experiments conducted on both procedurally generated scenarios and real-world scenarios show that increasing the diversity and the size of the training set leads to the improvement of the RL agent's generalizability. We further evaluate various safe reinforcement learning and multi-agent reinforcement learning algorithms in MetaDrive environments and provide the benchmarks. Source code, documentation, and demo video are available at \url{ https://metadriverse.github.io/metadrive}.
\end{abstract}

}

\maketitle
\IEEEdisplaynontitleabstractindextext
\IEEEpeerreviewmaketitle

\IEEEraisesectionheading{\section{Introduction}}
\IEEEPARstart{G}{reat} progress has been made in reinforcement learning (RL), ranging from super-human Go playing~\cite{silver2016mastering} to delicate dexterous in-hand manipulation~\cite{andrychowicz2020learning}. Recent progress shows the promise of applying RL to real-world applications, such as autonomous driving~\cite{kendall2019learning, sun2021neuro, https://doi.org/10.48550/arxiv.1912.00191}, the power system optimization in smart building~\cite{mason2019review}, the surgical robotics arm~\cite{richter2019open}, and even nuclear fusion~\cite{degrave2022magnetic}. However, generalization remains one of the fundamental challenges in RL. Even for the common driving task, an agent that has learned to drive in one town often fails to drive in another town~\cite{Dosovitskiy17}. There is the critical issue of model overfitting due to the lack of diversity in the existing RL environments~\cite{cobbe2019quantifying}. Many ongoing efforts have been made to increase the diversity of the data produced by the simulators, such as procedural generation in gaming environments~\cite{cobbe2019procgen} and domain randomization in indoor navigation \cite{li2021igibson}. 
In the context of autonomous driving (AD) research, many realistic driving simulators~\cite{Dosovitskiy17,martinez2017beyond,serban2019chrono,cai2020summit,zhou2020smarts} have been developed with their respective successes. 
Though these simulators address many essential challenges in AD, such as the realistic rendering of the surroundings in CARLA~\cite{Dosovitskiy17} and the scalable multi-agent simulation in SMARTS~\cite{zhou2020smarts}, they do not successfully address the aforementioned generalization problem in RL, especially the generalization across different scenarios.
Since the existing simulators mostly adopt fixed assets such as hand-crafted traffic driving rules and maps, the scenarios available for training and testing are far from enough to catch up with the complexity of the real world.

\begin{figure*}[!t]
  \centering
  \includegraphics[width=0.98\linewidth]{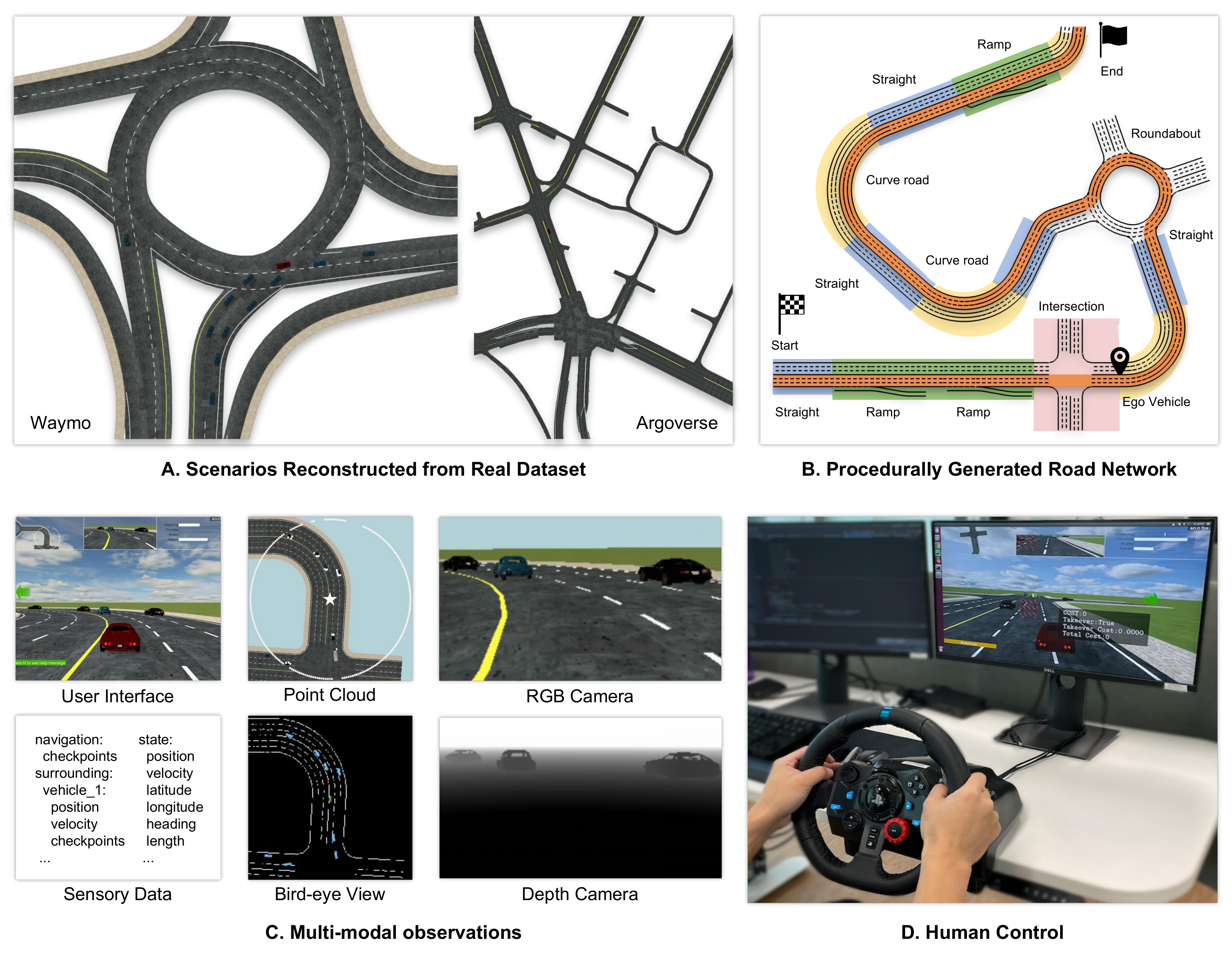}
  \caption{
  \textbf{A.} MetaDrive supports importing maps and traffic flow from real-world dataset.
  \textbf{B.} A road map procedurally generated from elementary road blocks. 
  \textbf{C.} Multi-modal observations provided by MetaDrive, including Lidar-like \revision{point clouds}, RGB / depth camera, bird-view semantic map, and scalar sensory data. 
  \textbf{D.} MetaDrive supports the control and demonstration from human subject.
  }
  \label{fig:teaser}
\end{figure*}

To better benchmark the generalizability of RL algorithms in the autonomous driving domain, we develop MetaDrive, a driving simulation platform that can compose a wide range of diverse traffic scenarios. 
MetaDrive holds the key feature of compositionality, inducing a diverse set of driving scenarios from procedural generation (PG), real data importing, and domain randomization. It also supports versatile research topics like safe RL and multi-agent RL.   
Every asset in MetaDrive, such as the vehicle, the obstacle, or the road structure, is defined as an interactive object with many configurable settings restricted in the parameter spaces, which can be combined, managed, and actuated to compose new interactive driving scenarios. 
As shown in Fig.\ref{fig:teaser}~\textbf{A.} and Fig.\ref{fig:teaser}~\textbf{B.}, through the procedural generation or the real data importing, interactive and executable scenarios can be generated to train and test learning-based driving systems.

Another challenge for generalizable RL research is the scalability issue~\cite{cobbe2019procgen}. Conducting experiments to benchmark the generalizability of RL agents requires massive computing resources and a long training time~\cite{cobbe2019quantifying}. 
To tackle this, MetaDrive incorporates a trade-off between the visual rendering and the physical simulation. 
With the resource saved from photorealistic rendering, a single MetaDrive instance with 100 MB size can run realistic physical simulation up to 300 FPS on a standard PC. The developed platform is also compatible with common RL training frameworks such as RLLib~\cite{liang2018rllib} and Stable-Baseline3~\cite{stable-baselines3}, which can launch more than 100 instances to generate batch data in parallel.   
Furthermore, MetaDrive can be installed with a single command line and interact with OpenAI Gym API in the Python environment. With those features, MetaDrive aims to facilitate the development of generalizable RL algorithms. 

MetaDrive supports a variety of reinforcement learning tasks because of its compositionality and extensibility. In the current development stage, we construct four standard RL tasks and implement their baselines in the context of driving research as below. The first three tasks are in single-agent setting while the last one is in multi-agent setting:
\begin{itemize}
  \item \textbf{Generalization to unseen PG scenarios}. Based on procedural generation, our simulator composes a large number of diverse driving scenarios from the elementary road structures and traffic vehicles. Those maps and scenarios are further split into training and test sets. we then conduct baseline experiments to evaluate the generalizability of different RL methods with respect to the road network structures and traffic flows.
  \item \textbf{Generalization to unseen real scenarios}. MetaDrive has the APIs to import map data collected in real-world such as Waymo dataset~\cite{waymo_open_dataset, sun2020scalability} and Argoverse dataset~\cite{chang2019argoverse}. Thus we can benchmark the generalizability of RL agents in real traffic scenarios. These real-world cases are closely examined and selected by humans to ensure the data fidelity, then they are divided into standard training and test set. Agents trained in scenes generated by PG can also be evaluated in real traffic cases to verify the effectiveness of PG, since the scenes are composed under a unified structure.
  \item \textbf{Safe exploration}. We study the safe exploration problem in RL, where the agent has to drive under safety constraints. Besides the surrounding  vehicles, obstacles like broken-down vehicles and traffic cones are randomly scattered on the road and a cost is yielded if a collision happens. We test several safe RL algorithms to measure their applicability to the safety-critical scenarios. 
  \item \textbf{Multi-agent traffic simulation}. In five typical traffic scenarios such as roundabout and intersection, we study the problem of multi-agent RL for dense traffic simulation. There are 20 to 40 agents in a scene, where each vehicle is actuated by a continuous neural control policy. Collective motions can emerge from the coordination of the agents.
\end{itemize}


\begin{table*}[!t]
\centering
\caption{
Comparison of representative driving simulators.
}
\label{tab:comparison}
\begin{tabular}{cccccccccc}
\toprule
Simulator & 
\begin{tabular}[c]{@{}c@{}}Unlimited\\ Scenarios \end{tabular} & 
\begin{tabular}[c]{@{}c@{}}Vehicle\\ Dynamics \end{tabular} &
\begin{tabular}[c]{@{}c@{}}Multi-agent\\ Support \end{tabular} & 
\begin{tabular}[c]{@{}c@{}}Custom\\ Map \end{tabular} & 
\begin{tabular}[c]{@{}c@{}}Lidar or\\ Camera \end{tabular} & 
\begin{tabular}[c]{@{}c@{}}Real Data\\ Importing \end{tabular} &
\begin{tabular}[c]{@{}c@{}}Realistic Visual \\ Rendering \end{tabular} &
\begin{tabular}[c]{@{}c@{}}In Active\\ Maintenance\end{tabular} &
\begin{tabular}[c]{@{}c@{}}Lightweight\end{tabular} \\
\toprule
CARLA~\cite{Dosovitskiy17}      & &\cmark &\cmark &\cmark&\cmark &\cmark & \cmark & \cmark &
\\ \midrule 
SUMMIT~\cite{cai2020summit}      & &\cmark &\cmark &\cmark&\cmark &\cmark & \cmark & \cmark &
\\ \midrule 
MACAD~\cite{palanisamy2020multi}      & &\cmark &\cmark &\cmark&\cmark &\cmark & \cmark & \cmark &
\\ \midrule 
GTA V~\cite{martinez2017beyond}      & &\cmark & &&\cmark &&\cmark& &
\\ \midrule
Highway-env~\cite{highway-env}     & & & && & &&\cmark
&\cmark 
\\ \midrule 
TORCS~\cite{wymann2000torcs}      & &\cmark & &\cmark&\cmark & &&
&\cmark
\\ \midrule
Flow~\cite{vinitsky2018benchmarks}      & & & &\cmark& &\cmark&&
& \cmark
\\\midrule 
CityFlow~\cite{zhang2019cityflow}      & & &\cmark &\cmark& &\cmark&&
& \cmark
\\\midrule 
Sim4CV~\cite{muller2018sim4cv}     & & \cmark& &&\cmark & &\cmark&
&
\\\midrule 
Duckietown~\cite{gym_duckietown} & & & && \cmark &\cmark &&\cmark &\cmark
\\ \midrule %
SMARTS~\cite{zhou2020smarts} & & \cmark &\cmark &\cmark& \cmark & &&\cmark
&\cmark
\\ \midrule %
AIRSIM~\cite{airsim2017fsr} & & \cmark & && \cmark & \cmark &\cmark& \cmark

\\ \midrule %
SUMO~\cite{SUMO2018} & & & &\cmark& & \cmark &&\cmark
&\cmark
\\ \midrule %
MADRaS~\cite{santara2021madras} & & \cmark & \cmark&\cmark& \cmark & &&&\cmark
\\ \midrule %
Udacity~\cite{udacity} & & \cmark & & \cmark& \cmark & &&
\\ \midrule %
DriverGym~\cite{kothari2021drivergym} & & & & & &\cmark & &\cmark&\cmark
\\ \midrule %
DeepDrive~\cite{deepdrive} & & \cmark & && \cmark & &\cmark& & \\ \midrule %

\begin{tabular}[c]{@{}c@{}}\textbf{MetaDrive}\end{tabular}  
&\cmark & \cmark &\cmark & \cmark & \cmark & \cmark &  &\cmark & \cmark  \\ \bottomrule
\end{tabular}
\end{table*}

\section{Related Work}
\textbf{RL Environments}.
A large amount of RL environments have been developed to benchmark the progress of different RL problems. 
Arcade Learning Environments \cite{bellemare13arcade} and the MuJoCo Simulator \cite{todorov2012mujoco} are widely studied in single-agent RL tasks, and in multi-agent RL the Particle World Environment \cite{mordatch2017emergence} and SMAC \cite{samvelyan2019starcraft} have become the standard testing grounds.
Meta-World \cite{yu2020meta} is developed for multi-task and meta reinforcement learning.
To apply RL algorithms in the real world, one has to consider their safety and robustness in previously unseen scenarios. Safety Gym \cite{safety_gym_Ray2019} and ProcGen \cite{cobbe2019procgen} are proposed to benchmark the safety and generalizability of RL algorithms, respectively.
Similar to ProcGen \cite{cobbe2019procgen} which uses procedural generation to generate distinct levels of video games, our MetaDrive simulator can also procedurally generate an infinite number of driving scenes.
As shown in the following sections, most of the aforementioned RL tasks can be implemented and studied in the proposed MetaDrive simulator because of its flexibility for extension and customization. 

\noindent\textbf{Driving Simulators}. 
Because of public safety concerns, it is difficult to train and test the driving policies in the physical world on a large scale. Therefore, simulators have been used extensively to prototype and validate autonomous driving research. Apart from MetaDrive, there are lots of existing driving simulators that support RL research. Table~\ref{tab:comparison} presents a comprehensive comparison between different simulators.
The simulators GTA V~\cite{martinez2017beyond}, Sim4CV~\cite{muller2018sim4cv}, AIRSIM~\cite{airsim2017fsr}, CARLA~\cite{Dosovitskiy17} and its derived project SUMMIT~\cite{cai2020summit}, MACAD~\cite{palanisamy2020multi} realistically preserve the delicate appearance and transition of the physical world, such as lighting conditions, weather system, even the time of a day shift. In contrast to the high-fidelity rendering of these simulators, MetaDrive trades off the visual appearance quality for its high sample efficiency and extensibility, in order to support a wide range of research topics such as safe RL and multi-agent autonomy. \revision{MetaDrive has realistic vehicle dynamics and kinematics where chassis, wheels and joints are built on constraints and actuated by Bullet physics engine. This feature distinguishes it from other simulators, such as Udacity~\cite{udacity}, TORCS \cite{wymann2000torcs}, DeepDrive~\cite{deepdrive}, Duckietown~\cite{gym_duckietown}, Highway-env ~\cite{highway-env} and DriverGym~\cite{kothari2021drivergym}, which only simulate vehicle with simple kinematic model, $e.g.$, Bicycle mode~\cite{polack2017kinematic}, and simplify the driving problem to a high-level discrete control task or provide simplistic driving scenarios.}
The aforementioned simulators are mainly designed for the single-agent scenario.
In the Multi-agent Reinforcement Learning(MARL) context, CityFlow~\cite{zhang2019cityflow} and FLOW~\cite{wu2017flow} are two macroscopic traffic simulators that are based on SUMO~\cite{SUMO2018} focusing on traffic flow simulation and hence are not suitable to investigate the detailed behaviors of each learning-based agent.
For MARL research, MACAD~\cite{palanisamy2020multi} and MADRaS~\cite{santara2021madras} are two traditional platforms which are built on CARLA~\cite{Dosovitskiy17} and TORCS~\cite{wymann2000torcs} respectively.
The newly developed SMARTS~\cite{zhou2020smarts} provides an excellent testbed for the interaction of RL agents and social vehicles in atomic traffic scenes. 
MetaDrive also designs MARL environments that cover complex scenarios such as tollgate and parking lot, with a dense population of 50+ agents.
Compared to the existing simulators, the proposed MetaDrive holds the key feature of compositionality and extensiblity for fostering new research opportunity in generalizable reinforcement learning.

\begin{figure*}[!t]
  \centering
  \includegraphics[width=0.99\linewidth]{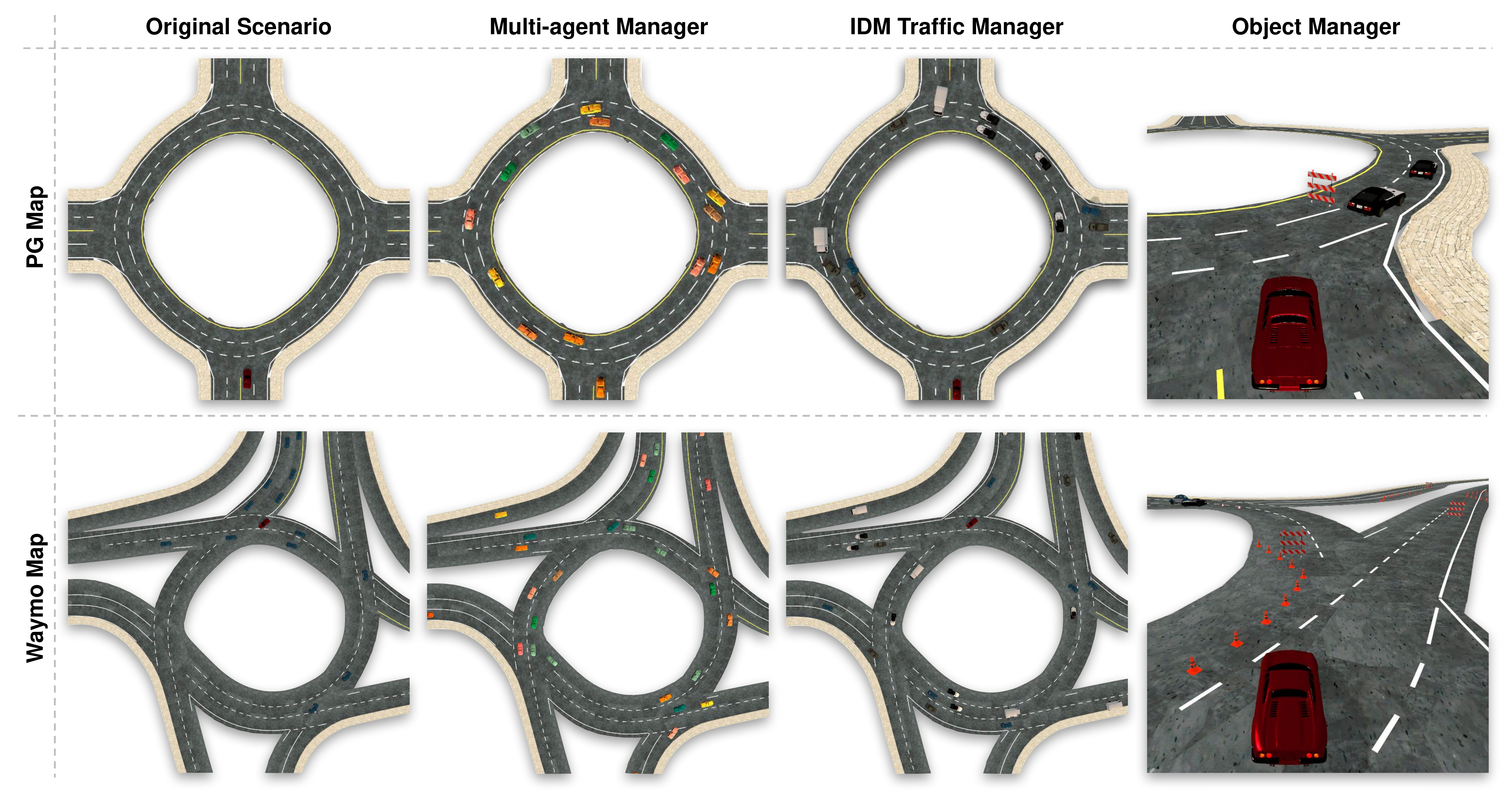}
  \caption{
  MetaDrive can compose new scenarios by combining various managers. 
  The first column shows two original cases from PG and Waymo datasets. 
  In the second column, by replacing the Single-agent Manager with the Multi-agent Manager, we compose the multi-agent navigation tasks in two maps. 
  In the third column, IDM Traffic Manager is added to both maps, enabling responsive traffic vehicles following IDM policies. The last column indicates that it also supports composing safety-critical environments on both PG map and real map by adding the Object Manager.
  }
  \label{fig:composition}
\end{figure*}

\section{System Design of MetaDrive}

The core feature of MetaDrive is its compositionality. 
This feature comes in two folds: the abstraction of low-level system implementation and the high-level aggregation of elementary components into traffic scenarios.
In low-level implementation, we encapsulate the back-end implementation of the basic components and the intercommunication between Python API and the extremely efficient 3D game engine Panda3D~\cite{goslin2004panda3d}.
With the high-level APIs, we implement various methods to generate traffic flows and diverse road networks.
The wrapping of back-end simulation makes it convenient and flexible for RL researchers to develop new scenarios and tasks to prototype their methods in the autonomous driving domain.
The diverse driving scenarios in MetaDrive also become the standard environments to benchmark various RL algorithms. In this section, we first introduce the key designs in MetaDrive and the abstraction of object classes. We then introduce the procedure of composing diverse scenarios from the elementary components.

\subsection{Core Concepts}

\noindent \textbf{Object}.
In MetaDrive, we abstract the object as the intermediate connector between the simulation engine and the Python environment.
Object is the basic entity in driving scenarios, such as vehicle, obstacle, traffic light, and road structure. In the back-end simulation, an object is a proxy to two internal models: the \textit{physical model} and the \textit{rendering model}.
Powered by the Bullet engine, the physical model in various shapes has a rigid body that can participate in collision detection and motion calculation. 
For retrieving realistic camera data, the rendering model provides fine-grained rendering effects such as light reflection, texturing, and shading. 

In the Python environment, the object class wraps the aforementioned details, handles the events such as garbage collection and exposes only the high-level APIs for manipulating the object and retrieving its information directly, such as \codefragment{object.set\_state()}, 
\codefragment{object.get\_state()}, 
\codefragment{object.step()}, 
and \codefragment{object.reset()}.
This design can not only achieve efficient and realistic simulation but enable users to manipulate objects and acquire their information with one line Python code.
Each class of objects has a parameter space that bounds the possible configurations of the instantiated objects, enabling randomizing and diversifying objects. For instance, a vehicle has parameters such as wheel friction, suspension stiffness, color, size, and so on. The parameters can be directly determined from user-specified configuration or randomly sampled in the parameter space. The object will automatically assign the determined parameters to the internal models.
With such design, a developer who wants to create and manipulate an object, such as setting the location and velocity of a vehicle, increasing the width of a lane, or getting nearby objects, can simply call the APIs without touching the complex back-end simulation system.

\begin{figure*}[!t]
  \centering
  \includegraphics[width=0.98\linewidth]{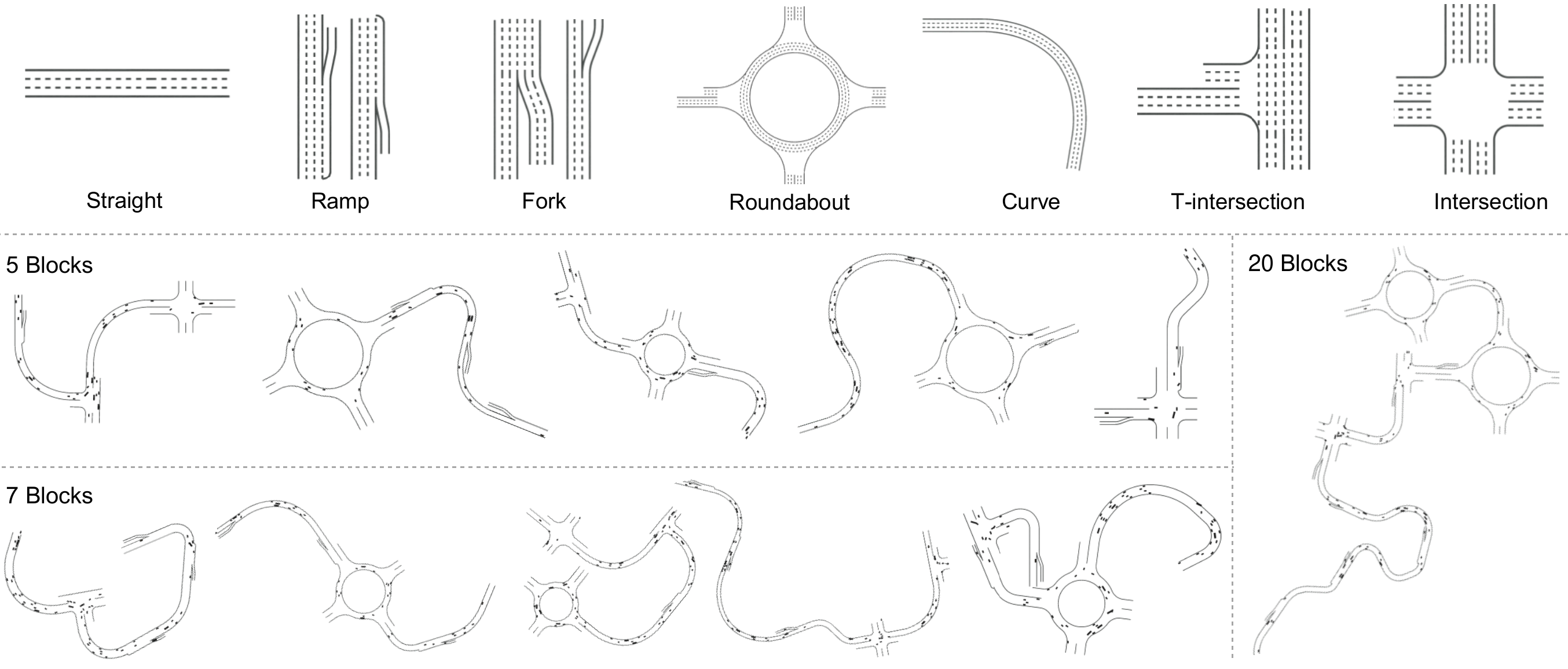}
  \centering\caption{First row shows the basic road blocks and the other rows plot the generated maps with different block numbers.}
  \label{fig:blocks_and_big_case}
\end{figure*}

\noindent\textbf{Policy}. 
A policy is a function that takes the object and environmental states as input and determines the action for controllable objects or new states for static objects. In each time step, policy of each object will be automatically called and the control signals, namely the actuation for vehicle or new position for obstacle, are passed to the objects by calling \codefragment{object.step()} or \codefragment{object.set\_state()} functions. The simulation engine then executes those control signals and advances the world for one step.

For vehicles, we currently implement a rule-based policy that mixes the cruising, lane changing, emergency stopping behaviors with various driving models such as IDM and mobile policy~\cite{kesting2007general}.
Furthermore, a well-trained PPO policy~\cite{schulman2017proximal} is encapsulated in the simulator, serving as another way to control traffic vehicles.
Additionally, MetaDrive provides policies that accept commands from external controllers such as neural policy and human control.
Besides the traffic participants, the road facilities can be also controlled by policies. 
For instance, in the MARL Tollgate environment Fig.~\ref{fig:marl-and-safety-environments}~\textbf{B.}, the tollgate follows \textit{Tollgate Policy} which releases the vehicle after it halts for a few seconds inside the gate.

The relationship between policies and objects is defined by users and can be varied during the episode, making it possible to do a mix-policy simulation. Similar to SMARTS~\cite{zhou2020smarts}, the mix-traffic simulation can be easily implemented by, for example, creating a scenario where some vehicles are controlled by neural policies, the other vehicles are actuated by IDM policies.

\noindent\textbf{Manager}. 
Different managers are used to manage objects in different roles. 
In MetaDrive, same class of objects might have different roles.
For example, in a single-agent RL environment, though the ego vehicle and the traffic vehicles are all vehicle objects, they have different roles and thus should have different policies. 
Different roles require various data processing pipelines, policies, and spawning/recycling rules. 
The ego vehicle requires the environment to provide explicit fine-grained surrounding information and is controlled by the RL agent. In contrast, the traffic vehicles rely on the rule-based policy and can be actuated with less detailed observations and states of other objects. Their spawning rules are also different. The ego vehicle is only created when an episode resets, while traffic vehicles will be recycled if they locate far from the ego vehicle. Therefore, the ego vehicle and the traffic vehicles should be managed by the \textit{Agent Manager} and the \textit{Traffic Manager} respectively.
A custom manager should implement \codefragment{manager.get\_state()} and \codefragment{manager.set\_state()} for retrieving and setting the states of managed objects, 
\codefragment{manager.step()} for invoking policies and actuating managed objects and \codefragment{manager.reset()} for resetting objects when meeting the termination conditions.

\subsection{Workflow for Scenario Composition}
Scenario composition in MetaDrive is conducted hierarchically: MetaDrive environment creates a set of managers according to user specification, then the managers spawn objects and assign policies to those objects if applicable.
After the initialization stage, all objects will run automatically in the environment following their policies while the managers monitor the states of the object and kick off new objects or recycle terminated ones. At each step, information and states of the managed objects will be thrown to the outside of the environment for policies, such as RL policies, to determine new control signals.

By flexibly combining existing managers, new environments, as well as the officially released benchmarks, can be easily composed for more intriguing research topics. For example, the \textit{Object Manager} in safe driving environments can be plugged into multi-agent environments with a simply \codefragment{engine.register\_manager()} and we get a new multi-agent environment for safe driving purposes. Furthermore, the \textit{Reply Traffic Manager} in real data replay environments can be replaced by \textit{IDM Traffic Manager} to generate reactive traffic on real maps instead of replaying logged traffic data.
By calling \codefragment{engine.update\_manager()}, as shown in Fig.~\ref{fig:marl-and-safety-environments}~\textbf{B.}, the \textit{Agent Manager} in generalization environments can be updated to \textit{Multi-agent Manager} to benchmark the generalizability of existing MARL algorithms. 
The combination of basic managers and their derived managers is thus called composition.
As shown in Fig.~\ref{fig:composition}, we provide some composition results.

\section{Composing Diverse Driving Scenarios}
In MetaDrive, a driving scenario can be composed by sequentially executing four basic Managers: 
(1) \textit{Map Manager}, in which PG maps or real maps will be randomly sampled for each new episode,
(2) \textit{Traffic Manager}, which contains a set of traffic vehicles cruising in the scene and navigating to their given destinations,
(3) \textit{Object Manager} that randomly scatters obstacles in the map to create more challenging and safety-critical scenarios,
(4) \textit{Agent Manager} that manages target vehicles which are actuated by external policies, $e.g.$ RL agents.
Other managers for special purposes, such as \textit{Tollgate Manager} and \textit{Traffic Light Manager}, are also available and can work with the four basic managers to compose more scenes. 


\subsection{Generating Maps}
MetaDrive supports two approaches to generate road networks: the procedural generation and importing from real data set.

\noindent\textbf{Procedural Generation}. One of the most important map sources is the Procedural Generation (PG).
As shown in Fig.~\ref{fig:blocks_and_big_case}, ingredients used to do PG are a set of road blocks sampled from typical types.
Each block preserves properties like lanes, spawn points for locating new vehicles, and \textit{sockets} for interconnect other blocks. 

\RestyleAlgo{ruled}
\begin{algorithm}[!t]
\caption{Procedural Generation of Maps}
\label{algo:pg}
\SetAlgoLined
\LinesNumbered
\DontPrintSemicolon
\KwIn{
Maximum tries in one block {\tt T}; Number of blocks in each map {\tt n}; Number of required maps {\tt N}\\
}
\KwResult{
 A set of maps $M =\{G^{(i)}_{net}\}_{i=1, ..., N}$
}
\SetKwProg{Fn}{Function}{}{}
\# Define the main function to generate a list of maps~\\
\Fn{{\tt \textbf{main(}T, n\textbf{)}}}{
  Initialize an empty list $M$ to store maps \\
   \While{$M$ does not contain {\tt N} maps}{
    Initialize an empty road network $G_{net}$ \\
    $G_{net}$, {\tt success}=\textbf{\textit{BIG({\tt T}, $G_{net}$, {\tt n})}} \\
    \uIf{{\tt success} is {\tt True}}{
      Append $G_{net}$ to $M$
    }
  }
\textbf{Return} {\tt M}
}
\nonl ~\\
\# Define the Block Incremental Generation (BIG) helper function that appends one block to current map if feasible and return current map with a success flag ~\\
\Fn{{\textbf{BIG(}{\tt T}, $G_{net}$, {\tt n}\textbf{)}}}{
  \uIf{$G_{net}$ has {\tt n} blocks}{
     \textbf{Return} {$G_{net}$, {\tt True}} 
  }%
  \For{{\tt 1, ..., T}}{
    Create new block $G_{\omega}$ = \textbf{\textit{GetNewBlock()}} \\
    Find the sockets for new block and old blocks: $e_1 \sim G_{\omega}(S)$, $e_2 \sim G_{net}(S)$ \\
    Rotate $G_{\omega}$ so that $e_1$, $e_2$ have supplementary heading\\
    \uIf{$G_{\omega}$ does not intersect with $G_{net}$}{
       $G_{net} \gets G_{net} \bigcup G_{\omega}$ \\
       $G_{net}$, {\tt success}=\textbf{\textit{BIG({\tt T},$G_{net}$, {\tt n})}} \\
      \uIf{{\tt success} is {\tt True}}{
         {\textbf{Return} { $G_{net}$, {True}}}
      }
      \uElse{
        Remove the last block from $G_{net}$
      }
    }
  }
  \textbf{Return} $G_{net}$, {\tt False}
}
\nonl ~\\
\# Randomly create a block ~\\
\Fn{{\textbf{GetNewBlock()}}}{
  Randomly choose a road block type $T \sim \{ \text{Staright}, ... \}$ \\
  Instantiate a block and randomize the parameters $G_{\omega}, \omega \sim \Omega_T$ \\
  \textbf{Return} $G_{\omega}$
}
\end{algorithm}

\begin{figure}[!t]
  \centering
  	\includegraphics[width=0.8\linewidth]{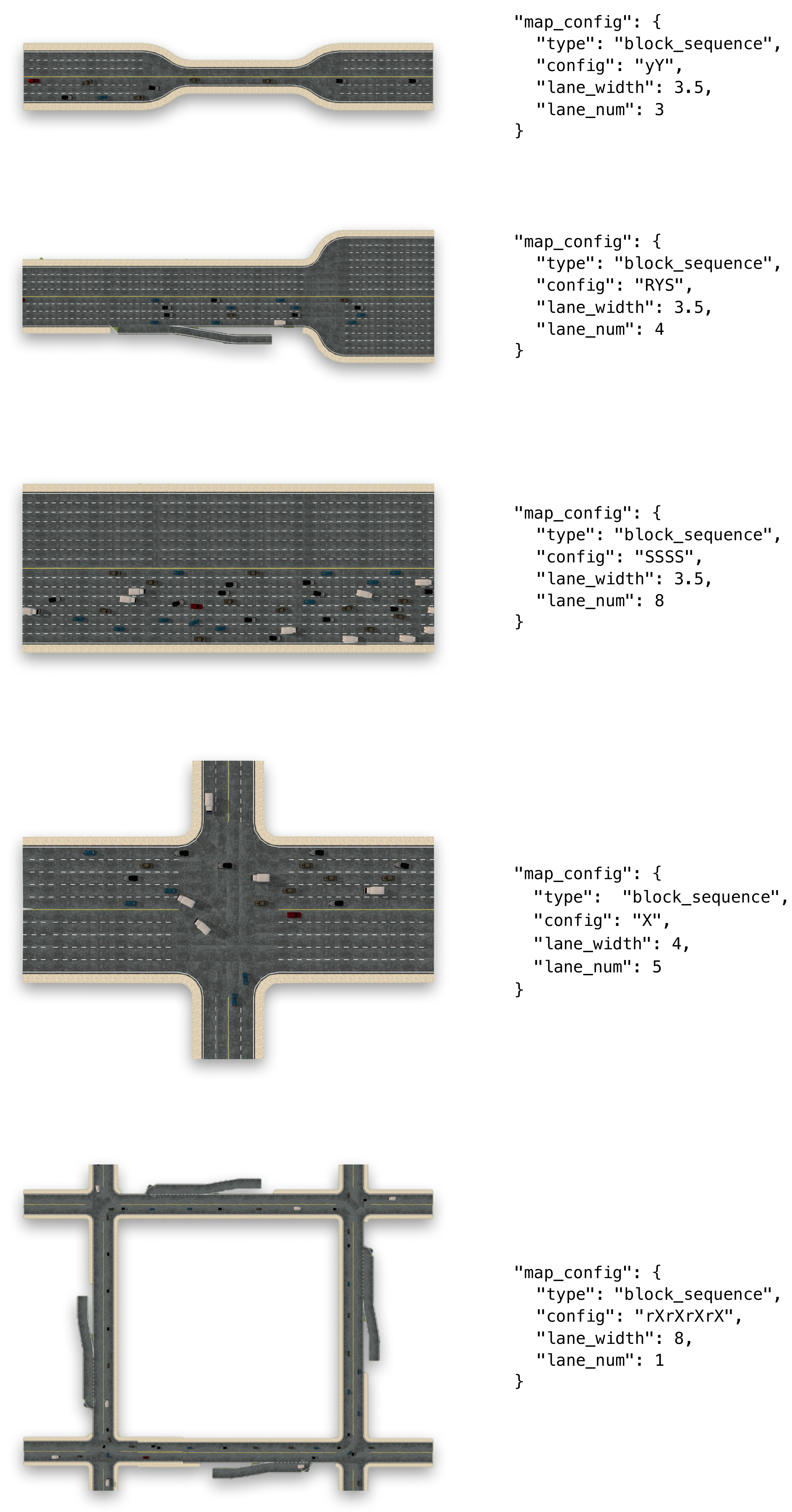}
  \caption{
  MetaDrive can derive diverse scenarios with different config in the input config. 
}
\label{fig:composition_show_case}
\end{figure}

\begin{figure*}[!t]
  \centering
  	\includegraphics[width=0.95\linewidth]{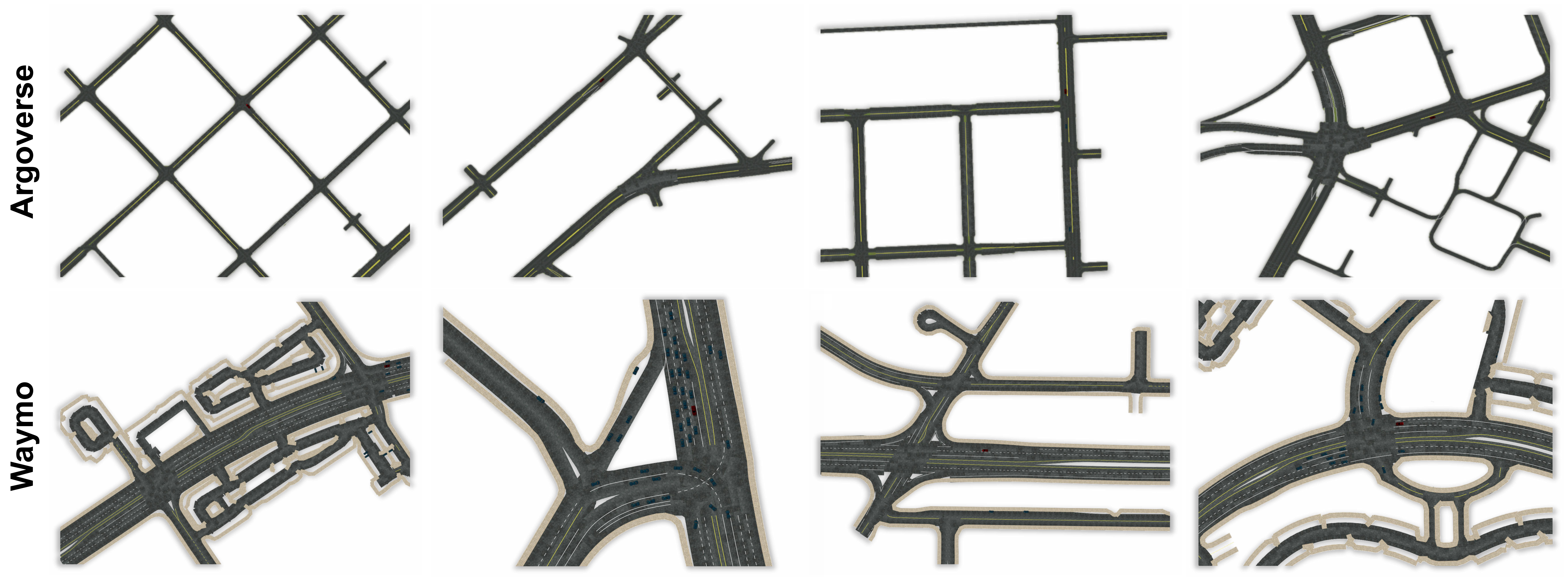}
  \caption{
  MetaDrive can load real scenarios from Argoverse dataset~\cite{chang2019argoverse} and Waymo dataset~\cite{sun2020scalability}.}
\label{fig:real_show_case}
\end{figure*}

\begin{figure}[!t]
  \centering
  \includegraphics[width=1\linewidth]{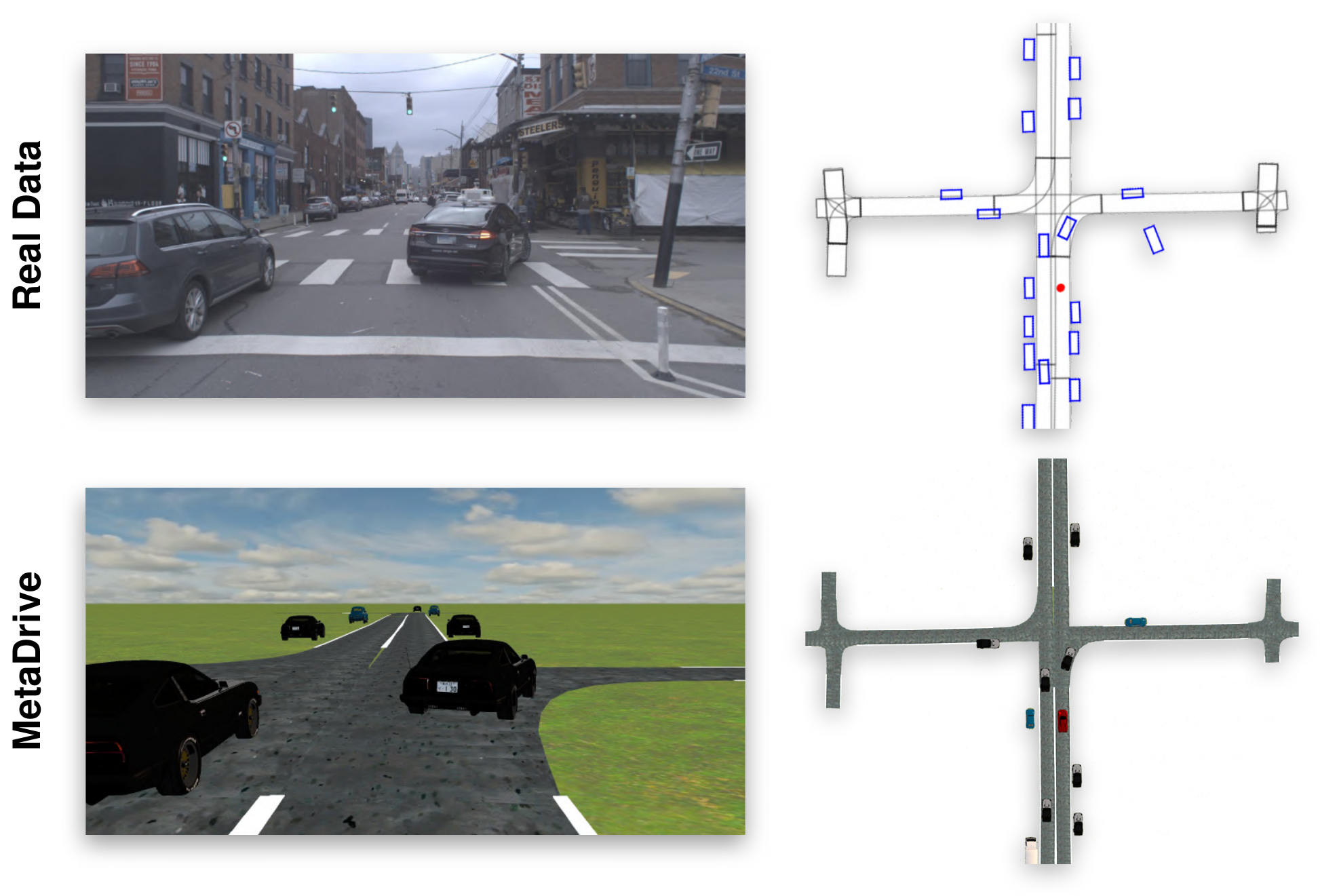}
  \vspace{-5mm}
  \caption{A simulation scenario is replicated from the real traffic data of Argoverse dataset~\cite{chang2019argoverse}.
  The traffic vehicles are actuated by Replay Traffic Manager.
  }
  \label{fig:argoverse_replay_showcase}
\end{figure}

We represent a road block using an undirected graph with additional features: $G_\omega=\{V, E, S, P, \omega, \Omega\}$, with nodes $V$ denoting the \textit{joints} in the road network and edges $E$ denoting \textit{lanes} which interconnect nodes. At least one node is assigned as the \textit{socket} $S=\{e_{i_1}, e_{i_2}, ...\}, e \in E$. 
The socket is usually a short straight road at the end of lanes which serves as the anchor to connect to the socket of other blocks. 
Block can preserve several sockets. For instance, Roundabout and T-Intersection have 4 sockets and 3 sockets, respectively.
Spawn points $P$ can be sampled on lanes for allocation of traffic participants. Apart from the above properties, a block type-specific \textit{parameter space} $\Omega$ is defined to bound the possible parameters $\omega$ of the block, such as the number of lanes, the lane width, and the road curvature, and so on.
The road block is the elementary component that can be assembled into a complete road network $G_{\omega_{net}} = \{ V, E, S, P, \omega_{net} \} = G_{\omega_1} \bigcup ... \bigcup G_{\omega_n}$ by the procedural generation algorithm.
Shown in Fig.~\ref{fig:blocks_and_big_case}, the detail of typical block type $T$ is summarized as follows:
\begin{itemize}
  \item \textbf{Straight}: A straight road is configured by the number of lanes, length, width, and lane line types.
  \item \textbf{Ramp}: A ramp is a road with an entry or exit existing in the rightest lane. The acceleration lane and deceleration lane are attached to the main road to guide the traffic vehicles to the enrty or exit of the main road.
  \item \textbf{Fork}: A structure used to merge or split additional lanes from the main road and change the number of lanes.
  \item \textbf{Roundabout}: A circular junction with four exits (sockets) with a configurable shape. Both roundabout, ramp, and fork aim to provide diverse merge scenarios.
  \item \textbf{Curve}: A curve block consists of a circular shape or clothoid shape lanes with configurable curvature.
  \item \textbf{T-Intersection}: An intersection that can enter and exit in three ways and thus has three sockets. The turning radius is configurable.
  \item \textbf{Intersection}: A four-way intersection allows bi-directional traffic. It is designed to support the research of unprotected intersection.
\end{itemize}
As illustrated in Algorithm~\ref{algo:pg}, we propose a search-based PG algorithm \textit{Block Incremental Generation (BIG)}, which recursively appends block to the existing road network if feasible and reverts the last block otherwise. 
When adding new block, BIG first uniformly chooses a roadblock type $T$ and instantiate a block $G_{\omega}$ with random parameters $\omega\sim \Omega_T$ (the function \textbf{\textit{GetNewBlock()}}). After rotating the new block so the new block's socket can dock into one socket of the existing network (Line 16, 17), BIG will then verify whether the new block intersects with existing blocks (Line 18). We test the crossover of all edges of $G_{\omega}$ and network $G_{net}$. If crossovers exist, then we discard $G_{\omega}$ and try a new one. Maximally {\tt T} trials will be conducted. If all of them fail, we remove the latest block and revert to the previous road network (Line 24).
The stop criterion of BIG is usually the number of blocks. In addition, users can also specify the block sequences to sample from the parameter spaces to generate maps with same block sequence but in different shapes. Fig.~\ref{fig:blocks_and_big_case} provides demonstration of generated maps assembled by different number of blocks.

\begin{figure*}[!t]
  \centering   
  \hfill%
  \includegraphics[width=0.95\linewidth]{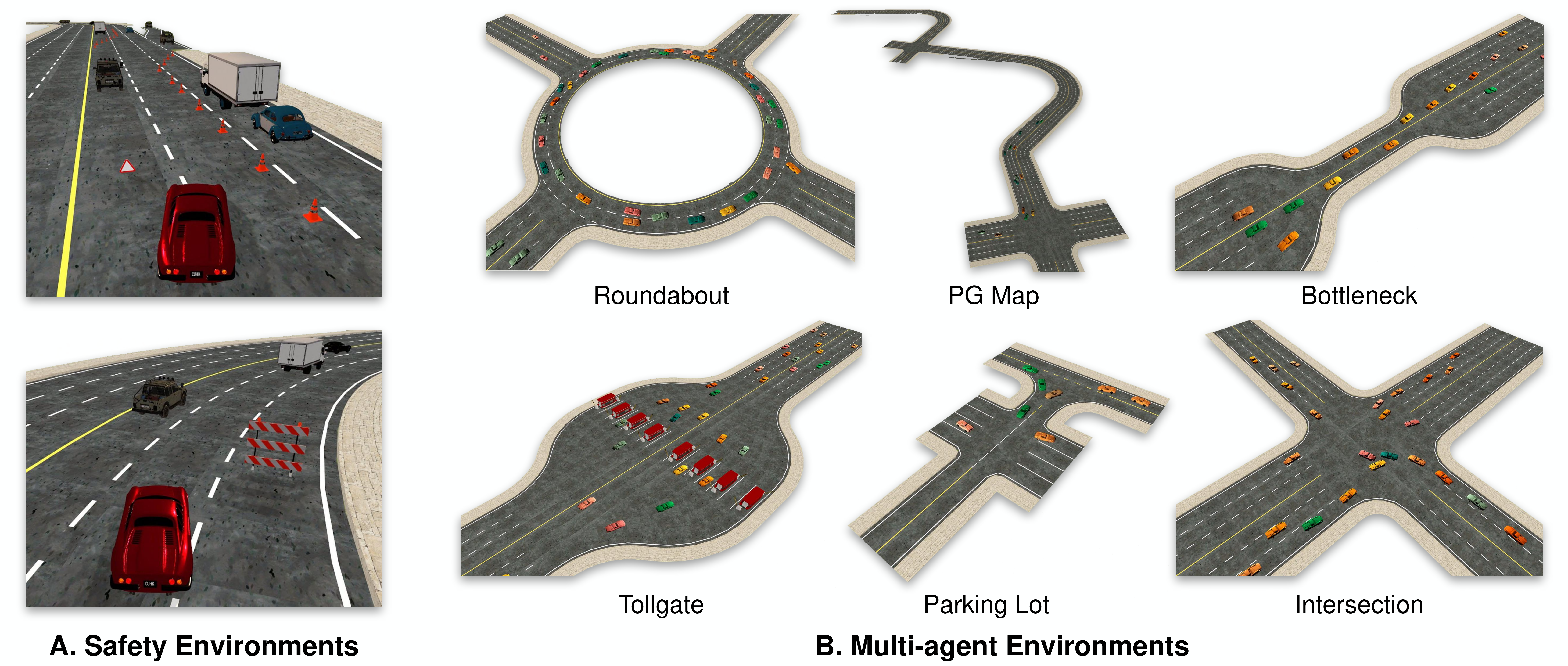}
 	\caption{
  \textbf{A.} Safe driving environments where obstacles such as broken down vehicles and traffic cones are randomly scattered on the map. \textbf{B.} Multi-agent environments where all agents need to coordinate their driving behaviors to achieve the population efficiency. 
  }
  \label{fig:marl-and-safety-environments}
\end{figure*}

The map generation process can be controlled by overwriting the map config, allowing the customization of the road network. When creating the environment, the user can pass a \codefragment{config}, namely a dictionary, into the environment \codefragment{MetaDriveEnv(config)} and specify the road network by providing overrides to the default settings of the map. For instance, as shown in Fig.~\ref{fig:composition_show_case}, the user can generate a variety of environments such as wide straight roads by modifying the dictionary with few lines of code.
In the showcases, we first specify the method to generate maps.
\codefragment{config["map\_config"]["type"] = "block\_sequence"} indicates generating map containing a given sequence of block types. Alternatively, setting type to \codefragment{block\_num} requests the BIG algorithms to produce a map containing given number of blocks while the type of blocks should be randomized.
Detailed annotations of different blocks is given in the documentation of the MetaDrive.

\noindent\textbf{Real Map}. Apart from the procedural generation of new scenarios, MetaDrive can also import real traffic data from autonomous driving datasets.
The road network data usually consists of a set of lane line center points, such as in Argoverse dataset ~\cite{chang2019argoverse}, Waymo dataset ~\cite{waymo_open_dataset, sun2020scalability} and OpenStreetMap~\cite{haklay2008openstreetmap}. Benefiting from the unified data structure to represent road networks, MetaDrive can seamlessly incorporate real-world data through creating lanes from waypoints in the dataset and then building functionalities at those lanes, such as the transformation between world coordinates and Frenet coordinates.
MetaDrive currently supports Waymo motion dataset~\cite{waymo_open_dataset,sun2020scalability} and Argoverse motion dataset~\cite{chang2019argoverse}, since they both contain the relationship between lanes, such as the entry lanes, existing lanes, left and right neighboring lanes.
We provide the Fig.~\ref{fig:real_show_case} to show the imported real maps from both datasets.
It is worth noticing that the Waymo dataset includes more diverse road structures and more detailed information such as boundaries than the Argoverse dataset. On the other hand, the Waymo dataset includes more complete scenarios (approximately 20,000) than the Argoverse dataset (approximately 100). Thus we only conduct real data generalization experiments on Waymo data.


\subsection{Generating Traffic Participants}
\textbf{Traffic Generation}.
MetaDrive maintains the traffic through \textit{Traffic Manager}.
The \textit{Traffic Manager} decides when and where to generate or recycle traffic vehicles and also assigns policies to vehicles. 
To support the compositionality, 
The attributions of created traffic vehicles, such as vehicle type, powertrain parameters, behaviors (aggressive or conservative), spawn points, and destinations, can be customized or randomized.

For PG maps, \textit{IDM Traffic Manager} is the default traffic manager which actuates traffic vehicles according to rule-based IDM policy, so they are responsive to the target vehicles. There are two modes to initialize the traffic flow in PG scenarios: Respawn mode and Trigger mode. 
Respawn mode is designed to maintain traffic flow density. 
In Respawn mode, \textit{Traffic Manager} assigns traffic vehicles to random spawn points on the map.
The traffic vehicles immediately start driving toward their destinations after spawning.
When a traffic vehicle terminates, it will be re-positioned to an available spawn point.
On the contrary, the Trigger mode traffic flow is designed to maximize the interaction between target vehicles and traffic vehicles.
The traffic vehicles stay still in the spawn points until the target agent enters the trigger zone.
Taking the Intersection as an example. Traffic vehicles inside the intersection will be triggered and start moving only when the target vehicle trespasses into the intersection.

\revision{For real maps imported from Argoverse~\cite{chang2019argoverse} and Waymo dataset~\cite{sun2020scalability, waymo_open_dataset}, \textit{IDM Traffic Manager} or \textit{Replay Traffic Manager} can be used. When steering vehicles by Replay Traffic Manager, in the log-reply mode, traffic vehicles strictly reproduce the trajectories collected in the real world without any reaction to the surroundings. As shown in Fig.~\ref{fig:argoverse_replay_showcase}, traffic vehicles with lop-replay policy replicate the behaviors recorded in Argoverse data set. When in the IDM control mode, the IDM controller actuates all the traffic vehicles whose initial states, \textit{i.e.} position are synchronized with the first frame of log data. After that, they are reactive to the ego car and surroundings and make decisions at every step. This, in turn, yields new trajectories different from the recorded data.
}

\noindent\textbf{Scattering Obstacles}. To compose near-accidental scenarios, Object Manager scatters many obstacles such as cones, crash barriers as well as broken down vehicles on the road, as shown in Fig.~\ref{fig:marl-and-safety-environments}~\textbf{A.} 
The density of the obstacles determines the difficulty of the task. A collision with the obstacle yields a cost for the ego vehicle, which can be used to train safe RL algorithms.

\noindent\textbf{Target Vehicles Management}. \textit{Agent Manager} is designed to register and maintain controllable vehicles in both single-agent and multi-agent environments.
It processes and packs information collected from the environment, such as the observations, rewards, and user-specified information with corresponding agent id, and then conveys them to external algorithms for training.
Our special implementation of \textit{Multi-agent Manager} makes it feasible to benchmark both common MARL setting and the mix-motive RL~\cite{jaques2018intrinsic,peng2021learning}, where the number of active agents is varying and new agents spawn immediately once old ones terminate.

\begin{table*}[!t]
\centering
\caption{Hyper-parameters of benchmarked methods.}
\label{table:all-hyper-parameters}
\begin{minipage}{0.26\linewidth}
\centering
\begin{tabular}[!t]{ll}
\toprule
SAC/SAC-Lag Parameters & Value \\ \midrule
Discount Factor $\gamma$  & 0.99 \\
$\tau$ for target network update & 0.005 \\
Learning Rate        & 0.0001 \\ 
Environmental horizon $T$ & 1500 \\
Steps before Learning start & 10,000\\
Buffer Size & 1,000,000 \\
Prioritized Buffer & True \\
Train Batch Size & 256 \\
Initial Alpha & 1.0 \\ 
Penalty Learning Rate & 0.01 \\
Cost Limit for SAC-Lag & 1 \\
\bottomrule
\end{tabular}
\end{minipage}
\hspace{16pt}
\begin{minipage}{0.26\linewidth}
\centering
\begin{tabular}[!t]{ll}
\toprule
PPO/PPO-Lag Parameters & Value \\ \midrule
Discount Factor $\gamma$  & 0.99 \\
KL Coefficient       & 0.2  \\
$\lambda$ for GAE~\cite{schulman2018highdimensional} & 0.95 \\
Number of SGD epochs  & 20   \\
Train Batch Size & 8000 \\
SGD mini batch size & 100 \\
Learning Rate        & 0.00005 \\ 
Clip Parameter $\epsilon$ & 0.2 \\
Penalty Learning Rate & 0.01 \\
CPO Target KL Divergence & 0.01\\
Cost Limit for PPO-Lag/CPO & 1 \\
\bottomrule
\end{tabular}
\end{minipage}
\hspace{10pt}
\begin{minipage}{0.36\linewidth}
\centering
\begin{tabular}[!t]{ll}
\toprule
IPPO/CCPPO Parameters       & Value \\ \midrule
KL Coefficient       & 1.0  \\
$\lambda$ for GAE~\cite{schulman2018highdimensional} & 0.95 \\
Discount Factor $\gamma$ & 0.99 \\
Environmental steps per training batch & 1024 \\
Number of SGD epochs $K_p$  & 5   \\
SGD mini batch size & 512 \\
Learning Rate        & 0.0003 \\ 
Environmental horizon $T$ & 1000 \\
Neighborhood radius $d_n$  & 40 m \\
Number of random seeds        & 8 \\  
Maximal environment steps & 1M \\
\bottomrule
\end{tabular}
\end{minipage}
\end{table*}

\begin{figure*}[!t]
\vspace{1em}
  \centering   
  \begin{minipage}{0.36\linewidth}
  \includegraphics[width=\linewidth]{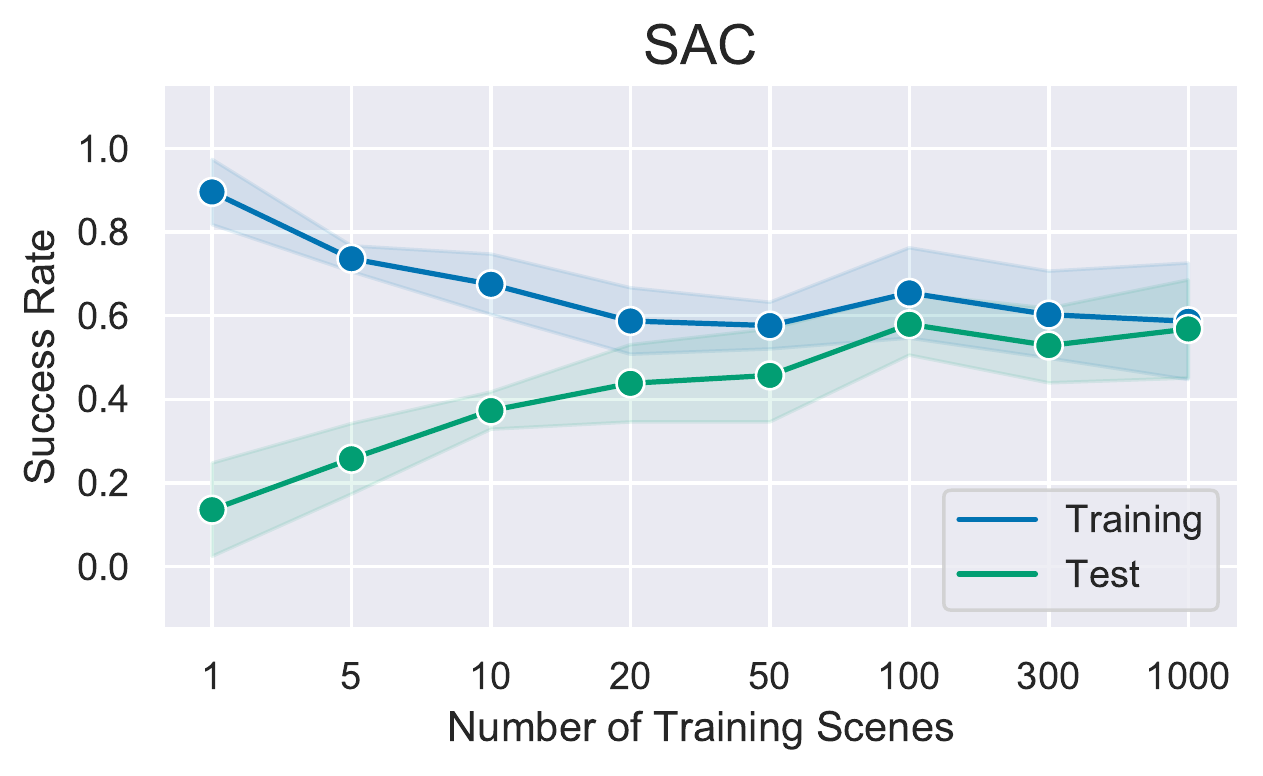}
  \end{minipage}%
  \begin{minipage}{0.36\linewidth}
  \includegraphics[width=\linewidth]{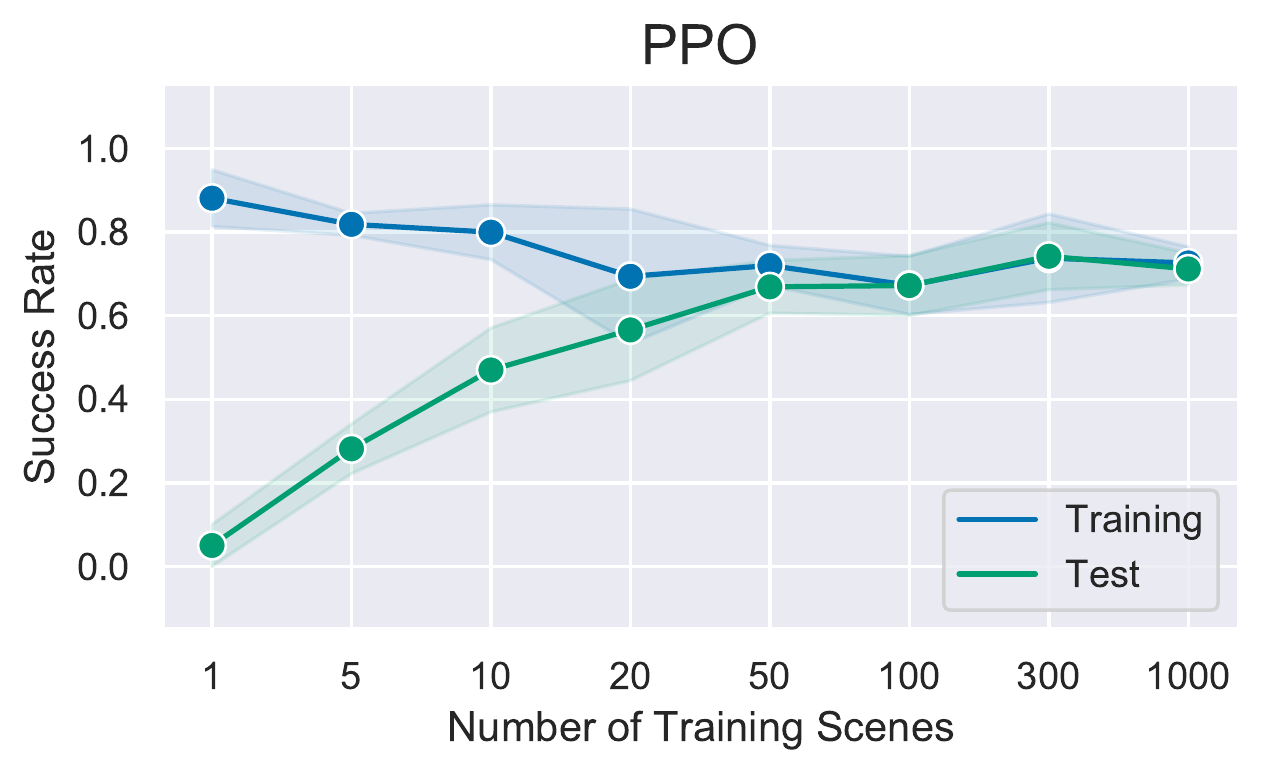}
  \end{minipage}%
  \begin{minipage}{0.24\linewidth}
  \includegraphics[width=\linewidth]{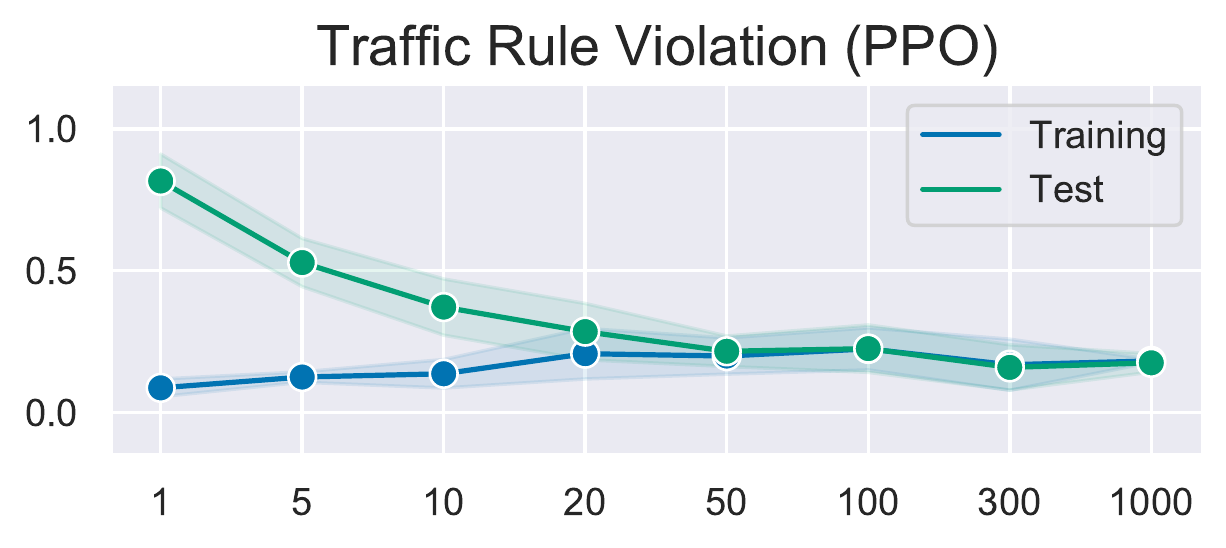}
  \includegraphics[width=\linewidth]{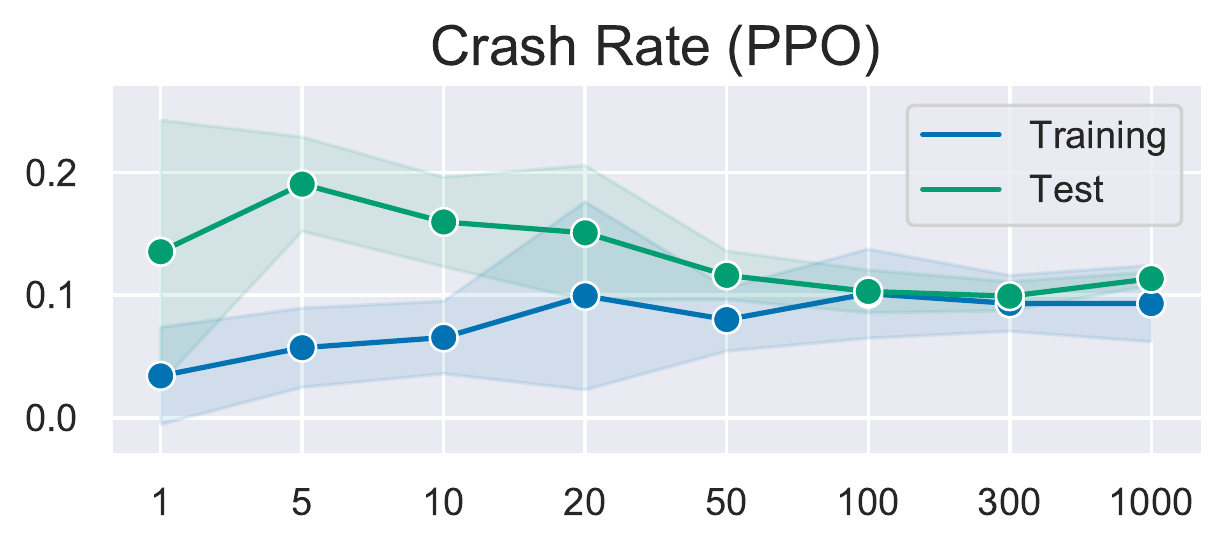}
  \end{minipage}%
  \hfill%
  	\caption{The generalization result of the agents trained with off-policy RL algorithm Soft Actor-critic (SAC)~\cite{haarnoja2018soft} and on-policy RL algorithm PPO~\cite{schulman2017proximal}.
  Increasing the number of training scenarios leads to a higher test success rate and lower traffic rule violation and crash probability, which indicates the agent's generalization is significantly improved.  
  Compared to PPO, the SAC algorithm brings a more stable training performance. The shadow of the curves indicates the standard deviation. 
  }
  \label{fig:main-result}
\end{figure*}

\subsection{
Sensors and Performance
}
\label{sect:implementation-details}
MetaDrive is implemented based on Panda3D~\cite{goslin2004panda3d} and Bullet Engine.
The well-designed rendering system of Panda3D enables MetaDrive to construct realistic monitoring and observational data. Bullet Engine empowers accurate and efficient physics simulation.
MetaDrive provides various kinds of sensory input, as illustrated in Fig.\ref{fig:teaser}~\textbf{C}.
For low-level sensors, Lidar, RGB cameras, depth cameras implemented by depth shader can be placed anywhere in the scene with adjustable parameters such as noise distribution, view field, and the laser number.
Meanwhile, MetaDrive can also provide high-level scene information as input to the learning policy, such as the road information like bending angle, length and direction, and nearby vehicles' information like velocity, heading, and profile.
Note that MetaDrive aims at providing an efficient platform to benchmark RL research, particularly generalizable RL research, therefore we improve the simulation efficiency at the cost of a photorealistic rendering effect. 
As a result, MetaDrive can run at 300 FPS in the single-agent environment with 10 rule-based traffic vehicles and 60 FPS in the multi-agent environment with 40 RL agents.
\section{Benchmarking Reinforcement Learning Tasks}
\label{section:experiments}
Based on MetaDrive, we construct four driving tasks corresponding to different reinforcement learning problems. 
The first two are in a single-agent setting where the traffic is actuated by rule-based policies (for PG maps) or trajectory replay policies (for real maps), and a target vehicle is controlled by an external RL agent.
The third task is in the safe driving setting where maps are generated by PG and obstacles are scattered randomly. 
The last task is in the multi-agent setting where a population of agents learns to simulate a traffic flow and each vehicle is actuated by a continuous control policy. We release the implementation of the baseline algorithms for reproducible research at \url{ https://github.com/metadriverse/metadrive}.

\subsection{Experimental Setting}
\label{section:experimental-setting}

In all tasks, the objective of RL agents is to steer the target vehicles with low-level continuous control actions, namely acceleration, brake, and steering.
We attempt to unify all tasks with a general setting of observation, reward function, and evaluation metrics.

\noindent\textbf{Observation}.
The observation of RL agents is as follows:
\begin{itemize}
  \item A 240-dimensional (72-dimensional in MARL) vector denoting the 2D-Lidar-like \revision{point clouds} with $50 m$ maximum detecting distance centering at the target vehicle. 
  Each entry is in $[0, 1]$ representing the relative distance of the nearest obstacle in the specified direction.
  \item A vector summarizing the target vehicle's state such as the steering, heading, velocity, and relative distance to the left and right boundaries.
  \item The navigation information that guides the target vehicle toward the destination. 
  We sparsely spread a set of checkpoints, \revision{50$m$ apart on average}, in the route and use the relative positions toward future checkpoints as observation to the target vehicle. 
\end{itemize}

\noindent\textbf{Reward and Cost Scheme}.
The reward function is composed of three parts as follows:
\begin{equation}
\label{eq:reward-functgion}
  R = c_{1}R_{disp} + c_{2}R_{speed} + R_{term}.
\end{equation}
The \textit{displacement reward} $R_{disp} = d_t - d_{t-1}$, wherein the $d_t$ and $d_{t-1}$ denote the longitudinal movement of the target vehicle in Frenet coordinates of current lane between two consecutive time steps, provides dense reward to encourage agent to move forward. 
The \textit{speed reward} $R_{speed} = v_t/v_{max}$ incentivizes agent to drive fast. $v_{t}$ and $v_{max}$ denote the current velocity and the maximum velocity ($80 \ km/h$), respectively.
We also define a sparse \textit{terminal reward} $R_{term}$, which is non-zero only at the last time step. At that step, we set $R_{disp} = R_{speed} = 0$ and assign $R_{term}$ according to the terminal state.
$R_{term}$ is set to $+10$ if the vehicle reaches the destination, $-5$ for crashing others or violating traffic rules such as locating at undrivable area.
We set $c_1 = 1$ and $c_2 = 0.1$. Sophisticated reward engineering may provide a better supervision signal, which we leave for future work.
For benchmarking Safe RL algorithms, collision to vehicles, obstacles, sidewalk and buildings raises a cost $+1$ at each time step.

\noindent\textbf{Evaluation Metrics}. We evaluate a given driving agent for multiple episodes and define the ratio of episodes where the agent arrives at the destination as the \textit{success rate}. The definition is the same for \textit{traffic rule violation rate} (namely driving out of the road) and the \textit{crash rate} (crashing other vehicles).
Compared to episodic reward, the success rate is a more suitable measurement when evaluating generalization, because we have a large number of scenes with different properties such as the road length and the traffic density, which makes the reward vary drastically across different scenes.
We conduct experiments on MetaDrive with algorithms mostly implemented in RLLib~\cite{liang2018rllib}. 
Each trial consumes 2 CPUs with 8 parallel rollout workers. 
All experiments are repeated 5 times with different random seeds.

\begin{figure*}
\centering
\includegraphics[width=\linewidth]{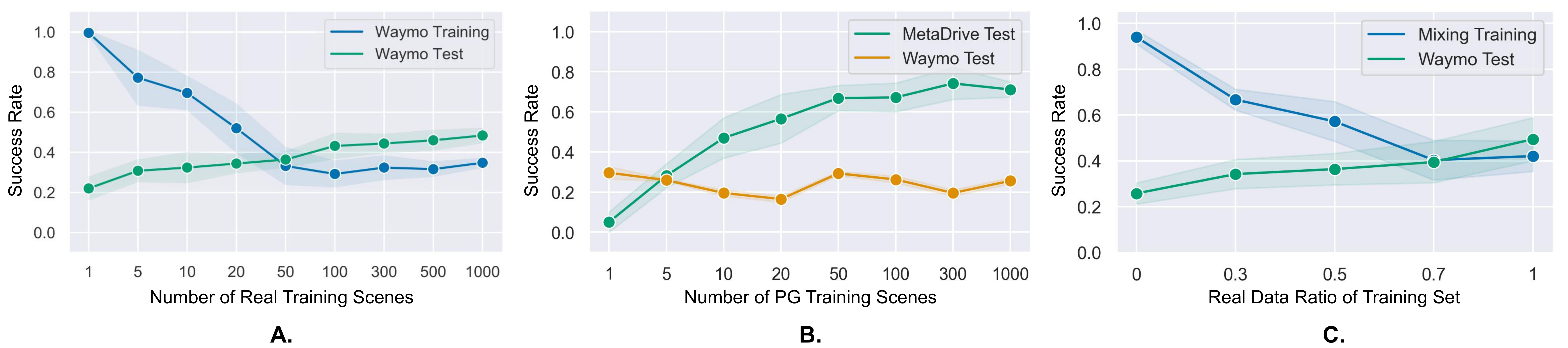}
\vspace{-2em}
\caption{
\textbf{A.} Waymo generalization experiment with a changing number of scenarios contained in training set.
\textbf{B.}
The test performance of PG-map-trained policies in real-world scenarios.
\textbf{C.} \revision{
Result of the policies trained on 5 training sets consisting of different mixing ratio of real and synthetic data.}
}
\label{fig:generalization_real_data}
\end{figure*}

\subsection{Generalization to Unseen PG Scenes}
\label{sect:pg-generalization}
To benchmark the generalizability of a driving policy, we develop an RL environment that can generate an unlimited number of diverse driving scenarios through the aforementioned procedural generation algorithm. Traffic flow powered by IDM model in varying density is also added to interact with the target vehicle.
We split the generated scenes into two sets: the training set and the test set. 
We train the RL agents only in the training set and evaluate them in the held-out test set.
The generalizability of a trained agent is therefore measured by the test performance.
The objective of this task is to show how the diversity and quantity of training scenarios affect the generalizability of the learned policy.
Table~\ref{table:all-hyper-parameters} describes detailed hyperparameter settings in this experiments.

We train the agents with two popular RL algorithms PPO~\cite{schulman2017proximal} and SAC~\cite{haarnoja2018soft}.
As shown in Fig.~\ref{fig:main-result}, the result of improved generalization capacity is observed in the agents trained from both RL algorithms: First, the overfitting happens if the agent is not trained with a sufficiently large training set.
When $N = 1$, namely the agent is trained in a single map, we can see the significant performance gap of the learned policies between the training set and test set.
Second, the generalization ability can be greatly strengthened if the agents are trained in more environments. As the number of training scenes $N$ increases, the final test success rate keeps increasing while rule violation and crash decrease drastically. 
The overfitting is alleviated and the test performance can match the training performance when $N$ is higher. 
The results clearly show that increasing the diversity of training environments can significantly improve the generalizability of RL agents. It also highlights the strength of the compositional simulator for generalizable reinforcement learning research. 

\subsection{Generalization to Unseen Real Scenarios}
MetaDive has the capacity to import real scenarios from autonomous driving dataset.
Currently, we build real-world cases collected by Waymo~\cite{waymo_open_dataset} and split the training set and test set so that they consist of 1000 and 100 non-overlapping cases, respectively. 

We conduct three experiments to discover the generalization phenomenon in real cases.
Firstly, by changing the number of cases contained in real training sets, we train 9 sets of agents to benchmark the generalizability of SAC~\cite{haarnoja2018soft}. 
As shown in Fig.~\ref{fig:generalization_real_data}~\textbf{A.}, increasing the training set size can also improve the generalizability of RL agents in real scenarios.
With the number of cases exceeding 100, the test performance is slightly higher than the training performance. 
This might because we build the test set with meticulous selection by human, while keeping the training scenarios draw from raw data. Therefore The noise and invalid cases in the training set undermine the training performance.

In the second experiment, we evaluate the trained PPO agents from PG generalization experiment (Sec.~\ref{sect:pg-generalization}) on the real-world test set.
The orange line in Fig.~\ref{fig:generalization_real_data} \textbf{B.} shows the test performance of those agents in Waymo dataset.
We find that increasing the diversity in PG training set can not improve the test performance on real-world test set.
This suggests the sim-to-real gap can not be eliminated by increasing training data from heterogeneous distribution.

The third experiment verifies whether introducing real cases in PG training set can improve the test performance in real test set.
We construct 5 training sets consisting of 100 training cases where real scenes and PG scenes are mixed in different ratios (0\%, 30\%, 50\%, 70\%, 100\%). 
As shown in Fig.~\ref{fig:generalization_real_data}~\textbf{C.}, increasing the real data ratio improves the test success on real test set. The decreasing training success rate implies that real scenarios is harder than PG scenes so that it is difficult for SAC to find a good solution. 

These three experiments show that for sim-to-real RL applications it is critical to ensure that data distributions in the simulator and real-world are similar. Otherwise, the generalizability shown in the simulation can not be well transferred to reality. 
By building training environments from real datasets, MetaDrive improves the generalization of trained policies in real-world scenarios and thus provides the flexibility to conduct further research.


\subsection{Safe Exploration}
As driving itself is a safety-critical application, it is important to evaluate the safe RL methods under the domain of autonomous driving. 
We define a new suite of environments to benchmark the \textit{safe exploration} in RL.
As shown in Fig.~\ref{fig:marl-and-safety-environments}~\textbf{A.}, we randomly place obstacles on the road. Different from the generalization task, we do not terminate the agent if a collision with obstacle or other vehicle happens. 
Instead, we allow the learning agent to continue driving and record the crash with a cost $+1$. Thus as a safe exploration task, the learning agent is required to balance the reward and the cost.
We also evaluate the agents on unseen maps to show their generalization ability in the aspect of safety.

\begin{figure}[!t]
  \centering   
  \begin{minipage}{\linewidth}
  \centering   
  \includegraphics[width=0.8\linewidth]{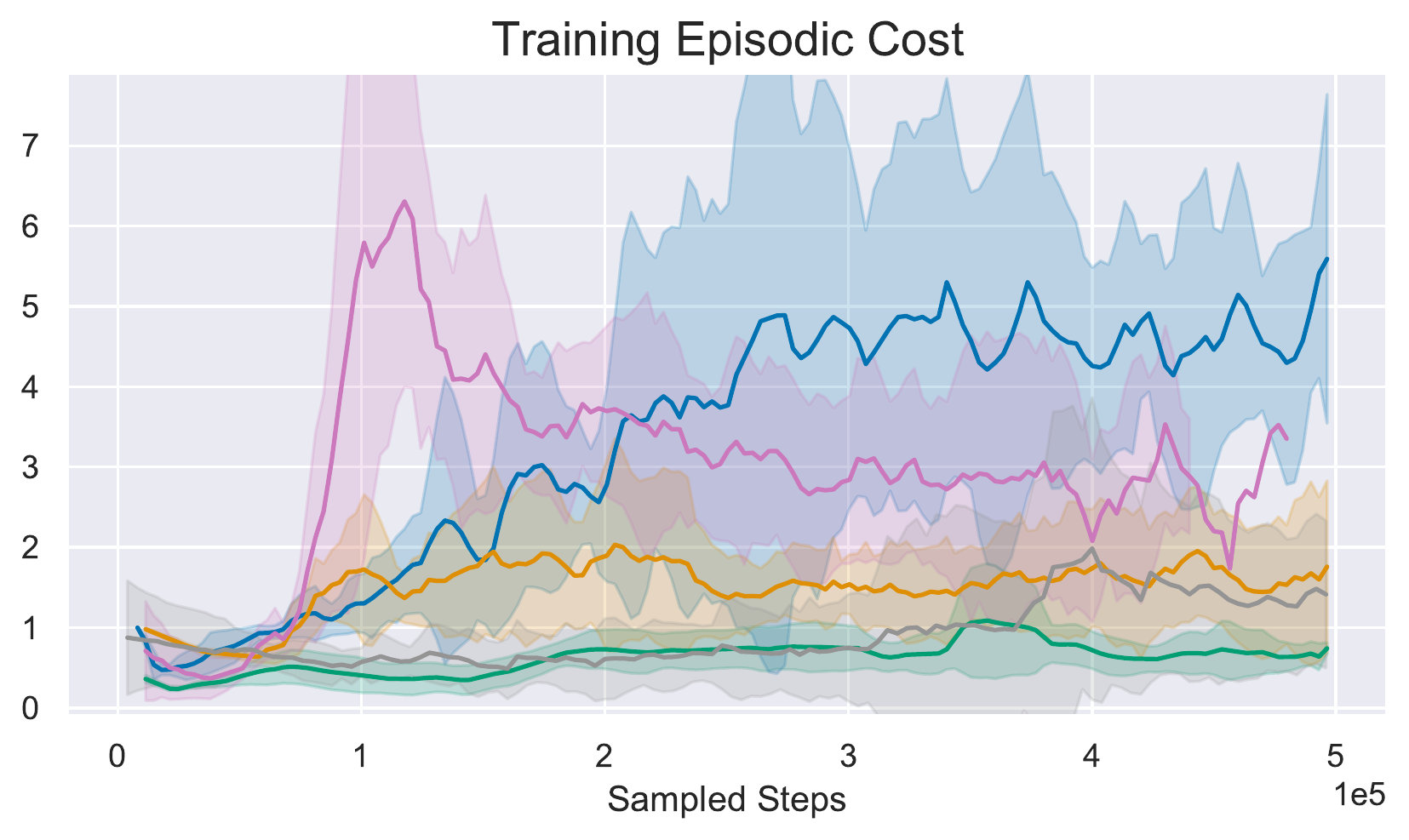}
  \end{minipage}\hfill%
  \begin{minipage}{\linewidth}
  \centering   
  \includegraphics[width=0.8\linewidth]{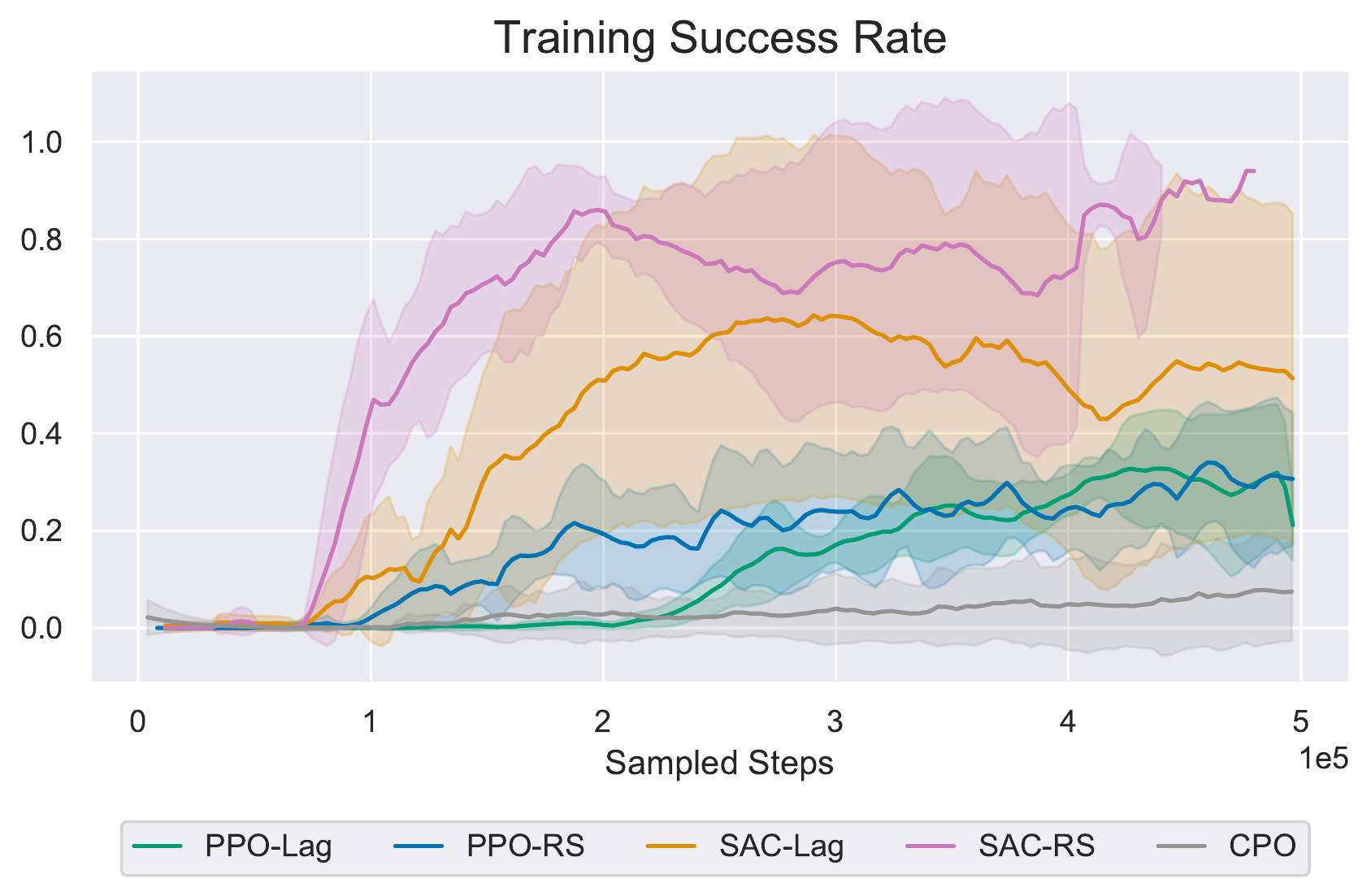}
  \end{minipage}%
  	\caption{Since the training time safety is critical to Safe RL, we show the learning progress of different Safe RL methods. Though achieves superior sample efficiency, the reward shaping version of SAC induces a peak in the training cost, while the Lagrangian SAC improves the policy while satisfying the safety constraint.
  	}
  \label{fig:safety-training-performance}
\end{figure}

We evaluate the reward shaping variants (RS) and Lagrangian variants (Lag) of PPO~\cite{schulman2017proximal} and SAC~\cite{haarnoja2018soft} as well as the Constrained Policy Optimization (CPO)~\cite{achiam2017constrained}. RS method considers negative cost as an auxiliary reward while Lagrangian method~\cite{safety_gym_Ray2019} consider the learning objective as:
\begin{equation}
\revision{
    \max_{\theta} \min_{\lambda \ge 0} = \mathop{E}_{\tau}[R_{\theta}(\tau) - \lambda( C_{\theta}(\tau)-d)],
    }
\end{equation}
wherein $R_{\theta}(\tau)$ and $C_{\theta}(\tau)$ are the episodic reward and the episodic cost respectively, $\theta$ is the policy parameters and $d$ is a given cost threshold. \revision{Thus the objective is to maximize episodic accumulative reward while restricting the episodic accumulative cost under a predefined threshold.}

Different from existing work which applies Lagrangian to SAC directly~\cite{ha2020learning}, we additionally equip SAC-Lag with a PID controller to update the multiplier to alleviate the oscillation in the training~\cite{stooke2020responsive}.
All Safe RL algorithms are trained using 500,000 steps. The algorithm configuration is in Table~\ref{table:all-hyper-parameters}. We also provide two offline learning baselines, BC and CQL~\cite{kumar2020conservative}, which use 36k human demonstrated transitions in 97\% success rate to do offline learning and are totally safe in training time. 
To promote imitation learning (IL) and Offline RL research, the driving trajectories collected by humans are also released at \url{https://github.com/metadriverse/metadrive/releases/download/MetaDrive-0.2.3/human_traj_100_new.json}.

As shown in Table~\ref{tab:safety-results} and Fig.~\ref{fig:safety-training-performance}, SAC-RS shows superior performance but causes high safety violations.
On the contrary, the Lagrangian SAC can achieve lower cumulative costs while giving up little reward.
Meanwhile, PPO is inferior compared to SAC, since it is an on-policy algorithm that are less sample-efficient than off-policy methods.
Fig.~\ref{fig:safety-training-performance} presents the learning dynamics of different Safe RL algorithms. The result suggests that SAC-RS shows the best sample efficiency, but a peak of episodic cost happens when it learns most efficiently. Noticeably, for BC and CQL, though interaction with environments is not required and policy can be learned with zero cost, they all have worse performance than online learning RL methods. CQL still outperforms BC, indicating that Offline RL method can learn a better policy given limited data.

\begin{figure}[!t]
  \centering
  \includegraphics[width=\linewidth]{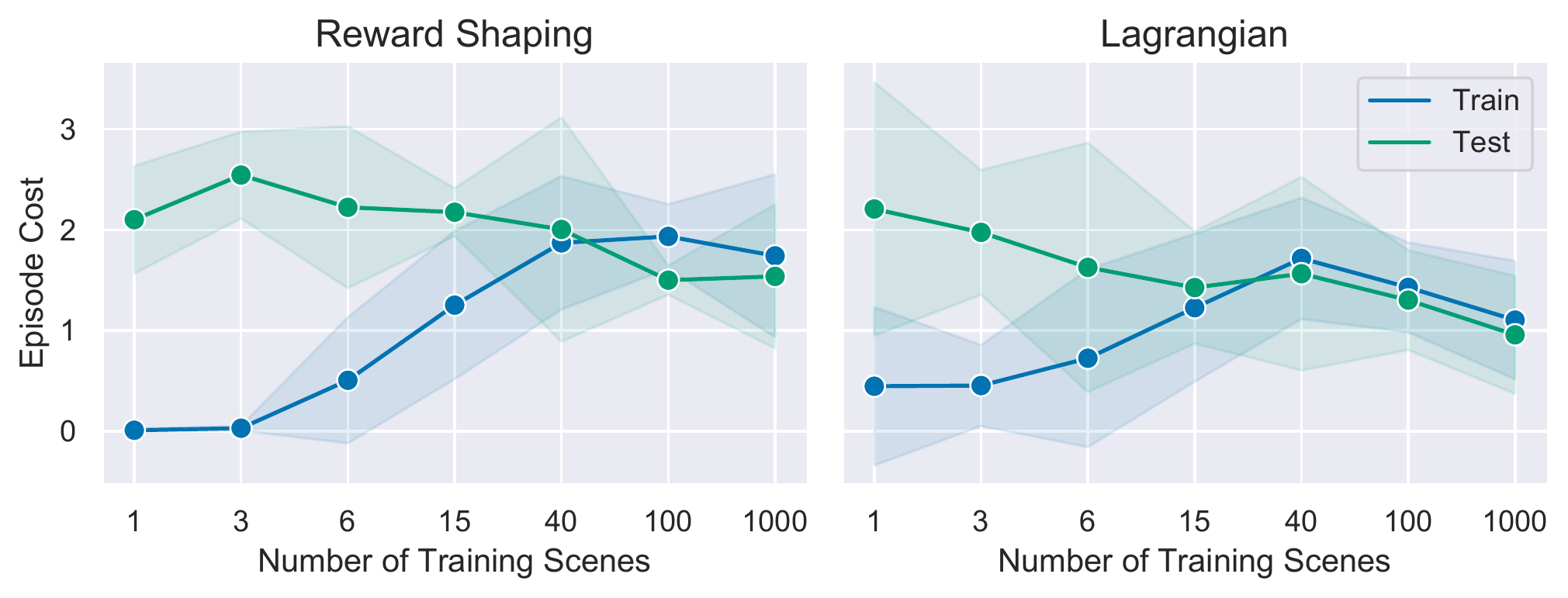}
  \caption{
  The episode cost of the trained policies in safety generalization experiment. We observe overfitting and poor safety performance in test time if trained with few training scenes. 
  }
  \label{fig:safety-result}
\end{figure}

We also conduct the \textit{safety generalization} experiment to verify the impact of training sets to testing-time safety performance.
We train the PPO-RS and PPO-Lag on training sets with different size and show the training and test episode cost in Fig.~\ref{fig:safety-result}.
When training with few environments, both methods achieve high test costs even if the training cost is low.
The safety generalizability can be improved by increasing the training set diversity. 
This experimental result implies the overfitting in safe exploration, which is a critical issue if we want to apply the end-to-end driving system to the real world.

\begin{table}[!t]
\centering
\caption{The test performance of different approaches in Safe Exploration.
}
\label{tab:safety-results}
\begin{tabular}{@{}ccccc@{}}
\toprule
Category & Method & ›
\begin{tabular}[c]{@{}c@{}}Cumulative \\ Reward \end{tabular}&
\begin{tabular}[c]{@{}c@{}}Cumulative \\ Cost \end{tabular} & 
\begin{tabular}[c]{@{}c@{}}Success Rate \end{tabular}\\
\toprule
\multirow{2}*{\shortstack{RL}} 
& SAC-RS~\cite{haarnoja2018soft} & \textbf{327.13} {\tiny $\pm$7.28} &	3.38 	{\tiny $\pm$0.60} &	\textbf{0.801} 	{\tiny $\pm$0.040} \\
& PPO-RS~\cite{schulman2017proximal} & 197.27 {\tiny $\pm$16.24} &	3.33 {\tiny $\pm$0.68} &	0.207 {\tiny $\pm$0.052}  \\ \midrule
\multirow{3}*{\shortstack{Safe RL}}
& SAC-Lag~\cite{stooke2020responsive} & 324.23 	{\tiny $\pm$14.45} & 1.90 {\tiny $\pm$0.44} & 0.714 {\tiny $\pm$0.103} \\
& PPO-Lag~\cite{safety_gym_Ray2019} & 269.51 {\tiny $\pm$22.54} & 1.82 {\tiny $\pm$0.33} & 0.477 {\tiny $\pm$0.114} \\
& CPO~\cite{achiam2017constrained} & 194.06 {\tiny $\pm$108.86} & \textbf{1.71} {\tiny $\pm$1.02} & 0.210 {\tiny $\pm$0.290} \\ \midrule
\multirow{2}*{\shortstack{Offline \\ Methods}}
& BC 	& 101.63	{\tiny $\pm$16.06} &		1.00 	{\tiny $\pm$0.45} &	0.01 	{\tiny $\pm$0.03} \\
& CQL~\cite{kumar2020conservative} & 156.4	{\tiny $\pm$31.94} &6.82	{\tiny $\pm$5.1} &	0.11 	{\tiny $\pm$0.07} \\
\bottomrule
\end{tabular}%
\end{table}

\subsection{Mixed Motive Multi-agent Reinforcement Learning}
We design five multi-agent RL environments in MetaDrive to benchmark the MARL algorithms.
Depending on the environments, 10 to 40 agents are simultaneously running in the environment. 
It is difficult to experiment with such dense multi-agent traffic in previous simulators due to poor efficiency.
In MetaDrive, we manage to achieve the efficiency of 60 FPS with 40 agents running simultaneously in a shared environment.
Fig.~\ref{fig:marl-and-safety-environments}~\textbf{B.} shows six environments for MARL, and their details are introduced below:
\begin{itemize}
  \item \textbf{Roundabout}: A four-way roundabout with two lanes. 40 vehicles spawn during environment reset. This environment includes merge and split junctions.

  \item \textbf{Intersection}: An unprotected four-way intersection allowing bi-directional traffic as well as U-turns. Negotiation and social behaviors are expected to solve this environment. We initialize 30 vehicles.

  \item \textbf{Tollgate}: Tollgate scene includes narrow roads to spawn agents and ample space in the middle with multiple tollgates. The tollgates become static obstacles where crashing is prohibited. We request agents to stop within tollgate for 3s. The agent will fail if they exit the tollgate before being allowed to pass. 40 vehicles are initialized. Complex behaviors such as deceleration and queuing are expected. Additional states such as whether the vehicle is in tollgate and whether the tollgate is blocked are given.

  \item \textbf{Bottleneck}: Complementary to Tollgate, Bottleneck contains a narrow bottleneck lane in the middle that forces the vehicles to yield to others. We initialize 20 vehicles in this scene.

  \item \textbf{Parking Lot}: A compact environment with 8 parking slots. Spawn points are scattered in both parking lots and on external roads. 10 vehicles spawn initially and need to navigate toward external roads or enter parking lots. In this environment, we allow agents to reverse their cars to spare space for others.
Good maneuvering and yielding are the keys to solving this task.

    \item \textbf{PG Map}: All vehicles in PG generated three roadblocks maps are controlled by RL agents.
\end{itemize}
These multi-agent environments pose a new challenge under the setting of mixed-motive RL, because each constituent agent in this traffic system is self-interested and the relationship between agents is changing.

\begin{table}[!t]
\centering
\caption{Success rate (\%) of different approaches in Multi-agent RL benchmarks.}
\label{tab:marl-benchmarks}
\begin{tabular}{@{}ccccccc@{}}
\toprule
Method  & Bottle. &                                   Tollgate &                               Inter. &                                Round. &                                Parking &                                    PG Map \\
\toprule
\shortstack{IPO\\~\cite{schroeder2020independent}}          &  \shortstack{ 74.18 \\ \tiny $\pm$ 15.87 } &  \shortstack{ 74.72 \\ \tiny $\pm$ 15.82 } &   \shortstack{ 73.93 \\ \tiny $\pm$ 9.18 } &  \shortstack{ 64.55 \\ \tiny $\pm$ 5.17 } &   \shortstack{ 20.90 \\ \tiny $\pm$ 5.70 } &  \shortstack{ 83.82 \\ \tiny $\pm$ 4.40 } \\ \midrule
\shortstack{MF-CCPPO\\~\cite{yang2018mean}}     &  \shortstack{ 65.27 \\ \tiny $\pm$ 17.06 } &  \shortstack{ 48.62 \\ \tiny $\pm$ 32.84 } &   \shortstack{ 71.58 \\ \tiny $\pm$ 7.79 } &  \shortstack{ 68.95 \\ \tiny $\pm$ 4.78 } &   \shortstack{ 14.42 \\ \tiny $\pm$ 4.92 } &  \shortstack{ 79.63 \\ \tiny $\pm$ 4.71 } \\ \midrule
\shortstack{Concat-\\CCPPO} &  \shortstack{ 51.93 \\ \tiny $\pm$ 17.03 } &  \shortstack{ 31.92 \\ \tiny $\pm$ 26.01 } &  \shortstack{ 62.67 \\ \tiny $\pm$ 12.46 } &  \shortstack{ 64.50 \\ \tiny $\pm$ 8.46 } &    \shortstack{ 4.73 \\ \tiny $\pm$ 3.83 } &  \shortstack{ 75.38 \\ \tiny $\pm$ 8.69 } \\ \midrule
\shortstack{CL\\~\cite{narvekar2020curriculum}}           &  \shortstack{ 66.45 \\ \tiny $\pm$ 10.50 } &  \shortstack{ 52.99 \\ \tiny $\pm$ 22.86 } &  \shortstack{ 67.90 \\ \tiny $\pm$ 15.00 } &  \shortstack{ 81.51 \\ \tiny $\pm$ 4.49 } &   \shortstack{ 16.17 \\ \tiny $\pm$ 8.94 } &  \shortstack{ 77.96 \\ \tiny $\pm$ 6.95 } \\ \midrule
\shortstack{CoPO\\~\cite{peng2021learning}}          &  \shortstack{ 73.39 \\ \tiny $\pm$ 16.65 } &  \shortstack{ 79.66 \\ \tiny $\pm$ 13.92 } &   \shortstack{ 77.79 \\ \tiny $\pm$ 2.95 } &  \shortstack{ 73.65 \\ \tiny $\pm$ 4.61 } &   \shortstack{ 20.98 \\ \tiny $\pm$ 3.55 } &  \shortstack{ 80.24 \\ \tiny $\pm$ 4.21 } \\ 
\bottomrule
\end{tabular}%
\end{table}

We benchmark several MARL algorithms: Independent policy optimization (IPPO) method~\cite{schroeder2020independent} which uses PPO~\cite{schulman2017proximal} as the individual learners and two centralized critic methods which encode the nearby agents' states into the input of value functions (centralized critics).
We test two variants of centralized critic PPO (CCPPO):
The first one is the Mean Field (MF) CCPPO, which averages the states of nearby vehicles within 10 meters and feeds the mean states to the value network~\cite{yang2018mean}.
The second variant concatenates the state of K nearest vehicles (K=4 in our experiment) as a long vector feeding as extra information to the value network~\cite{pal2020emergent}. 
Recently proposed CoPO~\cite{peng2021learning} is also compared, which is a mixed-motive MARL method addressing the coordination problem in multi-agent systems.
The hyperparameters of used methods can be found in Table~\ref{table:all-hyper-parameters}.
Table~\ref{tab:marl-benchmarks} shows the MARL experiment results.
Note that the result is based on MetaDrive 0.2.5 and is different to previous paper~\cite{peng2021learning} due to updates in MetaDrive. 
We find that the concatenated state of K nearest vehicles hurts the performance of CCPPO and leads to worse performance compared to MF-CCPPO.
We hypothesize this is because in driving tasks the neighborhood of ego vehicles is varying all the time while concatenating states greatly expands the input dimension thus creating difficulty to the learning.
The CoPO method~\cite{peng2021learning} achieves good performance in various environments, due to the local and global coordination.

\subsection{Discussion on the Learning-based Driving Policies}
When developing MetaDrive, we ran numerous experiments to search proper driving policies, and found that the performance of learning-based methods greatly depends on the observation. RL methods learn faster using the state vector observation in Sec.~\ref{section:experimental-setting} than the first-person view RGB images or top-down semantic map. Also, we find the displacement reward is the most important part of the reward function. We hope these environmental setting can provide insights on designing end-to-end driving policy.

We have implemented and benchmarked many learning algorithms for machine autonomy, including Reinforcement Learning, Imitation Learning (IL), Offline RL methods. 
The experimental results show that 1) Value-based methods (SAC, TD3) is more suitable for driving tasks than policy-based method (PPO, TRPO) due to their high sample efficiency and superior success rate (Fig.~\ref{fig:main-result}, Table~\ref{tab:safety-results}); 2) For learning safety, it is important to develop new paradigm incorporating safety constraint into the policy instead of manually shaping reward and cost function. Our recent works on human-in-the-loop training~\cite{peng2021safe,li2022efficient} provide examples; 3)When learning from limited offline data, Offline RL outperforms IL by a large margin in the safe exploration experiment~\ref{tab:safety-results}. Also, in our previous work~\cite{peng2021safe}, we train CQL~\cite{kumar2020conservative} and BC with a large dataset containing 300,000 transitions collected by a well-trained PPO expert policy. The results show that the CQL can achieve 72\% success rate, while BC only reaches 13\%, suggesting that Offline RL is an appealing direction for future works.


\section{Ethical Statement and Limitations}

\noindent\textbf{Ethical Statement}. MetaDrive provides a driving simulation environment for RL research.
Though MetaDrive itself has marginal negative societal impact, there might be two possible cases that cause damages. First, if a user applies the trained agent from MetaDrive to a real vehicle, the vehicle may cause accidents due to domain gap or uncertainty in neural network/action distribution. There is still a long way to go to narrow the sim2real gap. 
Second, since MetaDrive allows human subject to control vehicle via keyboard, joystick, or steering wheel, the player may experience, though the chance is low, discomfort or pain due to exposure to the fast driving in the MetaDrive 3D visualization.

\noindent\textbf{Limitations}. MetaDrive has limitations in the following aspects: 
First, the image rendering is not realistic as other driving simulators with photorealistic rendering effects like CARLA~\cite{Dosovitskiy17}.
This is because MetaDrive specifically focuses on the generalizable RL research and thus aims to flexibly compose driving scenarios.
Second, the generated driving scenes from MetaDrive don't contain traffic participants such as pedestrians and bicyclists. 
Third, a systematic driving scenario description protocol is left for future work to enable large-scale generation of corner cases such as near-accidental scenes for prototyping safe autonomous driving systems~\cite{rss}.


\section{Conclusion}

We develop a compositional and extensible driving simulator MetaDrive to facilitate the research of generalizable reinforcement learning. MetaDrive holds the core feature of compositionality, where an infinite number of diverse driving scenarios can be composed through both the procedural generation and the real traffic data replay. 
We construct a variety of RL tasks and baselines in both single-agent and multi-agent settings, including benchmarking generalizability across unseen scenes, safe exploration, and simulating multi-agent traffic.

\noindent \textbf{Licenses}.
MetaDrive is released under Apache License 2.0. 
Panda3D, used in MetaDrive as a rendering system for visualization, is under the Modified BSD license, which is a free software license with very few restrictions on usage.
Bullet engine is under Zlib license.
The vehicle models are collected from Sketchfab under CC BY 4.0 or CC BY-NC 4.0.
The skybox images are under CC0 1.0 Universal (CC0 1.0).
To support real data import, 
MetaDrive is made using the Waymo Open Dataset~\cite{waymo_open_dataset,sun2020scalability}, provided by Waymo LLC under license terms available at~\url{https://waymo.com/open}. 
Also, Argoverse is provided free of charge under the Creative Commons Attribution-NonCommercial-ShareAlike 4.0 International Public license. Argoverse code and APIs are provided under the MIT license. \\\textbf{Acknowledgement:} This project was partially supported by the Centre for Perceptual and Interactive Intelligence (CPII) Ltd under the Innovation and Technology Fund.


\bibliography{root}
\bibliographystyle{plain}

%
\vspace{-1em}
\begin{IEEEbiography}[{\includegraphics[width=1in,height=1.25in,clip,keepaspectratio]{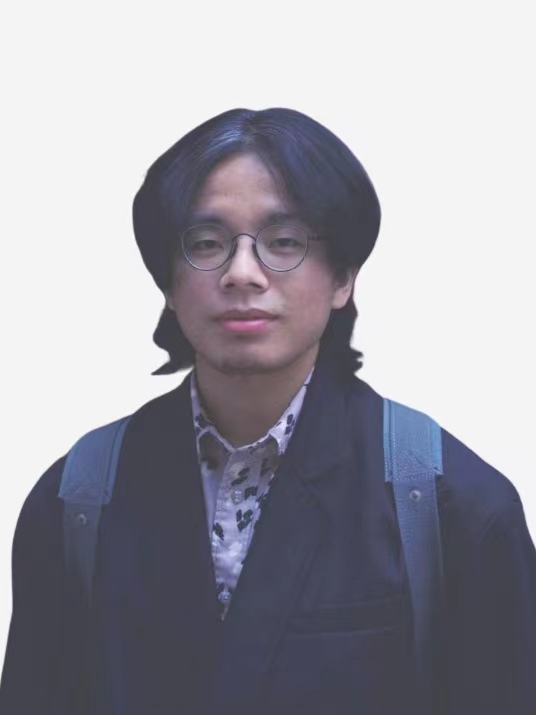}}]{Quanyi Li}
is a research assistant at Centre for Perceptual and Interactive Intelligence (CPII), the Chinese University of Hong Kong (CUHK). He received the B.Eng. degree in communication engineering from Beijing University of Posts and Telecommunications. He is interested in embodied AI and its generalizability, safety, interactivity and interpretability. 
\end{IEEEbiography}
\vspace{-3em}
\begin{IEEEbiography}[{\includegraphics[width=1in,height=1.25in,clip,keepaspectratio]{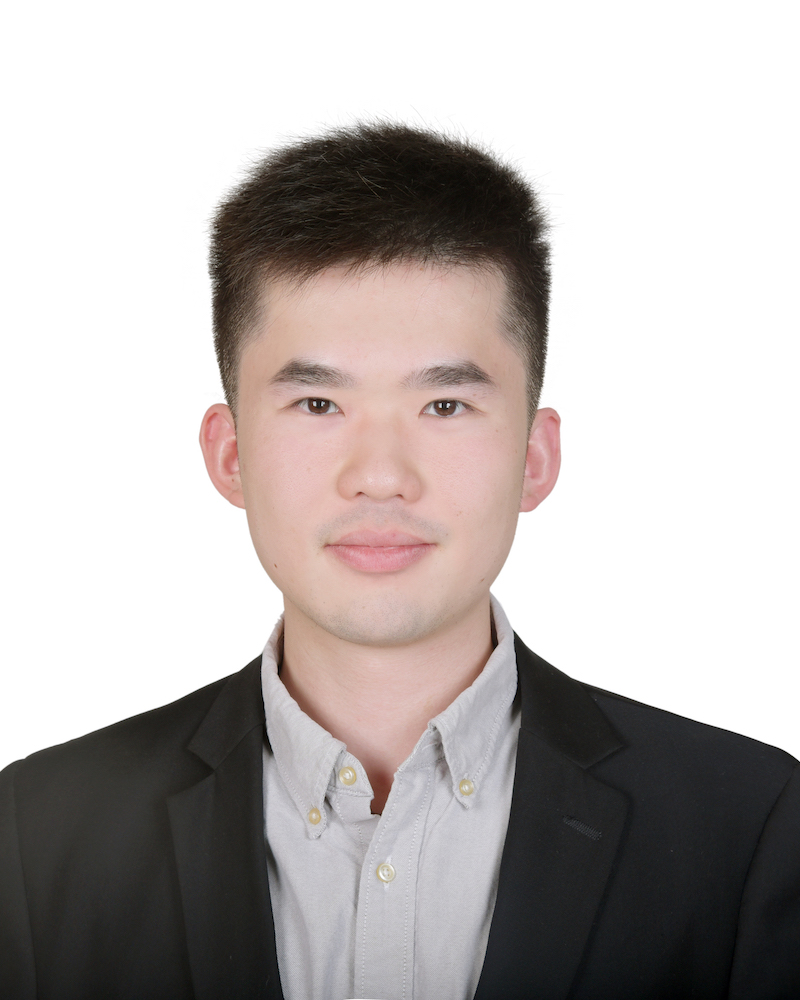}}]{Zhenghao Peng}
is a Ph.D. Candidate in Department of Computer Science at University of California, Los Angeles (UCLA).
He received M.Phil. degree in Department of Information Engineering at the Chinese University of Hong Kong (CUHK) and B.Eng. at Shanghai Jiao Tong university. His research interests are multi-agent reinforcement learning and human-AI interaction.
This work is done while he was at CUHK.
\end{IEEEbiography}
\vspace{-2em}
\begin{IEEEbiography}[{\includegraphics[width=1in,height=1.25in,clip,keepaspectratio]{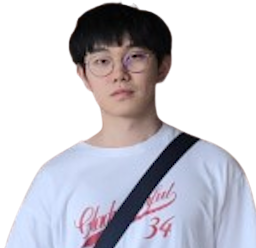}}]{Lan Feng}
is a master student majoring in robotics, system, and control at Eidgenössische Technische Hochschule Zürich (ETH Zurich). He received B.Eng. degree in Wuhan University (WHU). His research focuses on AI's generalizability and safety.
\end{IEEEbiography}
\vspace{-3em}
\begin{IEEEbiography}[{\includegraphics[width=1in,height=1.25in,clip,keepaspectratio]{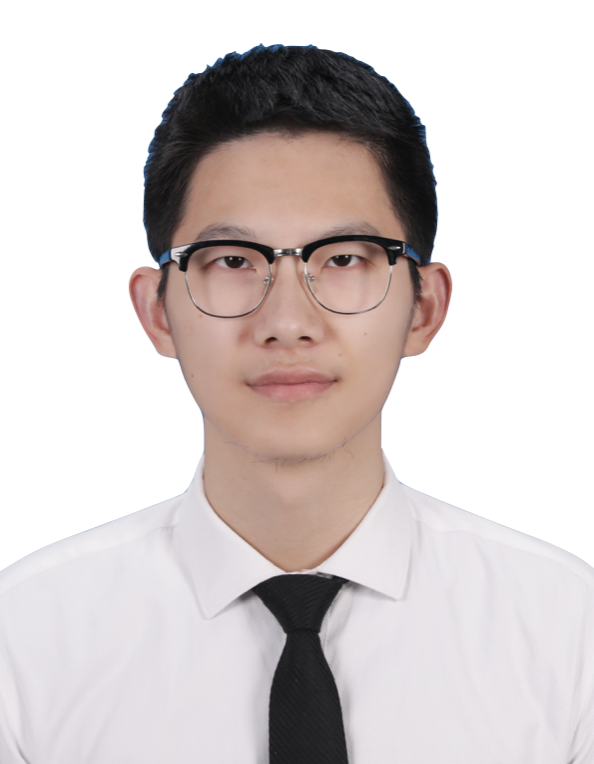}}]{Qihang Zhang}
is a Ph.D. Candidate in Department of Information Engineering at the Chinese University of Hong Kong (CUHK). He received the B.Eng. degree in computer science and technology from Zhejiang University (ZJU) in 2021. His research interest lies in embodied intelligence and machine autonomy and learning.
\end{IEEEbiography}
\vskip 0pt plus -1fil
\begin{IEEEbiography}[{\includegraphics[width=1in,height=1.25in,clip,keepaspectratio]{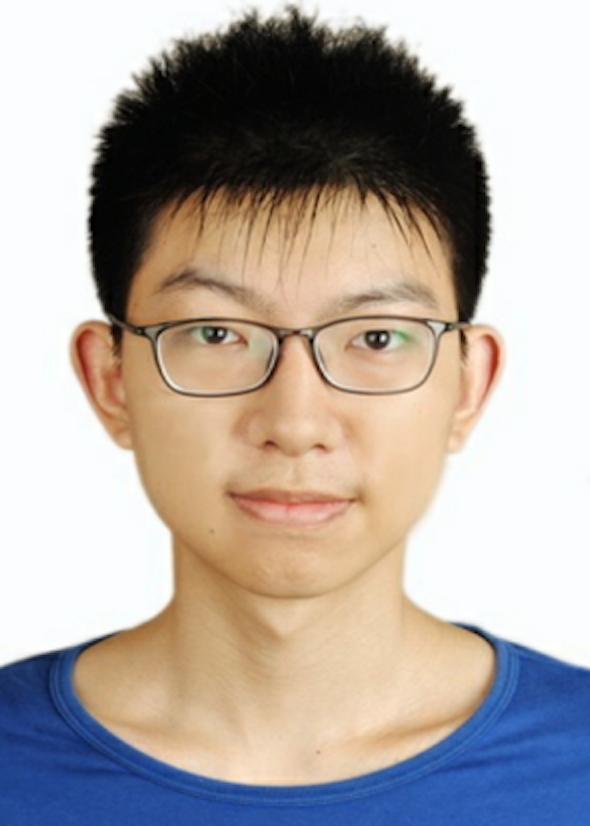}}]{Zhenghai Xue} is a PhD student in the School of Computer Science and Engineering, Nanyang Technological University (NTU) Singapore. He received a B.S. in the School of Artificial Intelligence, Nanjing University (NJU) in 2022. He worked as a research assistant in the Department of Information Engineering at the Chinese University of Hong Kong (CUHK) in 2021. His research interest is reinforcement learning.
\end{IEEEbiography}
\vskip 0pt plus -1fil
\vspace{-1em}
\begin{IEEEbiography}[{\includegraphics[width=1in,height=1.25in,clip,keepaspectratio]{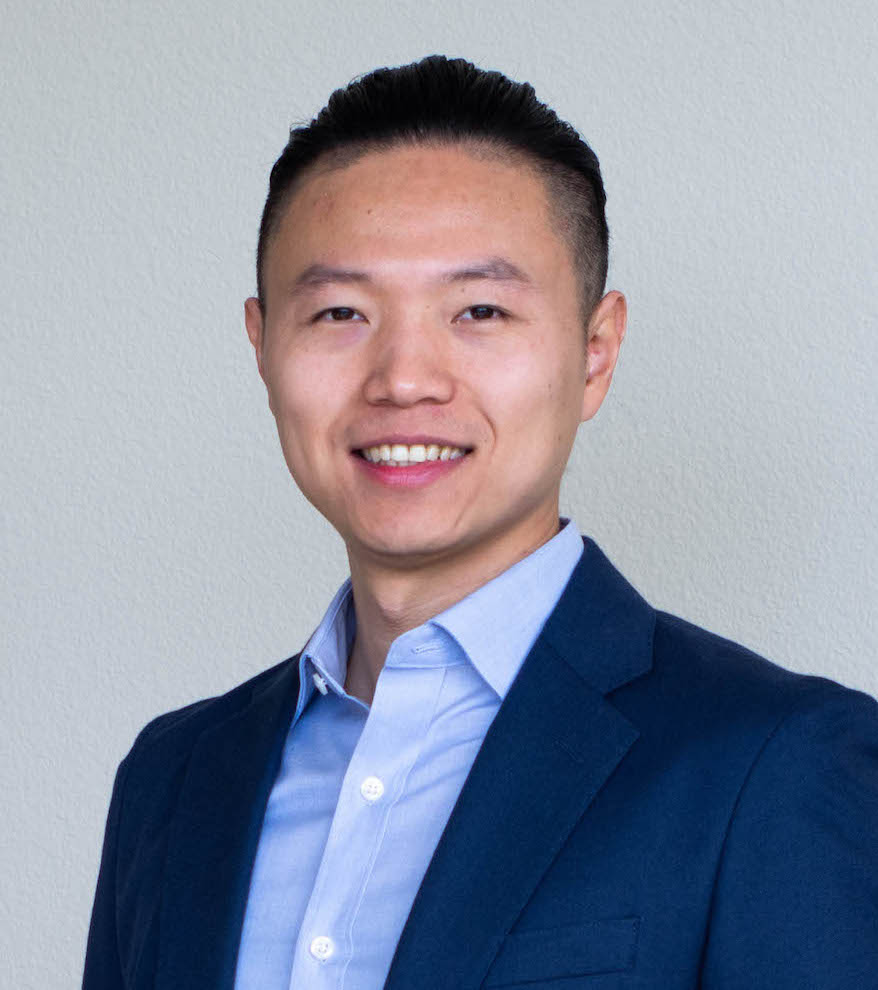}}]{Bolei Zhou} is an Assistant Professor in the Computer Science Department at the University of California, Los Angeles (UCLA). He earned his Ph.D. from MIT in 2018. His research interest lies at the intersection of computer vision and machine autonomy, focusing on enabling interpretable human-AI interaction. He and his colleagues have developed a number of widely used interpretation methods such as CAM and Network Dissection, as well as computer vision benchmarks Places and ADE20K. He is an associate editor for Pattern Recognition and has been area chair for CVPR, ICCV, ECCV, and AAAI. He received MIT Tech Review's Innovators under 35 in Asia-Pacific Award.
\end{IEEEbiography}


\end{document}